\documentclass[sigconf,screen]{acmart}
\setcopyright{none}
\acmPrice{15.00}
\acmDOI{}
\acmYear{2020}
\copyrightyear{2020}
\acmSubmissionID{issta20main-id50-p}
\acmISBN{978-1-4503-8008-9/20/07}
\acmConference[ISSTA '20]{Proceedings of the 29th ACM SIGSOFT International Symposium  on  Software Testing and Analysis}{July 18--22, 2020}{Los Angeles, CA, USA}
\acmBooktitle{Proceedings of the 29th ACM SIGSOFT International Symposium  on  Software Testing and Analysis (ISSTA '20), July 18--22, 2020, Los Angeles, CA, USA}

\usepackage{graphicx}

\usepackage{wrapfig}

\usepackage{caption}

\usepackage{paralist}

\usepackage{amsmath,amssymb,amsfonts}
\usepackage{algorithmic}
\usepackage{textcomp}
\usepackage{xcolor}

\usepackage{amsthm}
\usepackage{mdframed}
\usepackage{centernot}
\usepackage{comment}
\usepackage{stmaryrd}

\newcommand{\Pp}{\mathcal{P}}
\newcommand{\sPp}{\sem{\Pp}}

\newcommand{\Oo}{\mathcal{O}}
\newcommand{\Dd}{\mathcal{D}}

\newcommand{\sem}[1]{ [ \! [ {#1}  ]  \! ]} 
\def\rmdef{\stackrel{\mbox{\rm {\tiny def}}}{=}} 

\newcommand\vx{\mathbf{x}}

\newcommand\vd[1]{d_{p}}
\newcommand{\set}[1]{\left\{ #1 \right\}}

\newcommand{\Nat}{\mathbb N}

\newcommand{\R}{\mathbb R}

\newcommand{\Rplus}{\R_{\geq 0}}
\newcommand{\toolname}{\textsc{DPFuzz}\xspace}
\usepackage{tikz}
\newcommand*\circled[1]{\tikz[baseline=(char.base)]{
            \node[shape=circle,draw,inner sep=1pt,minimum size=0.1cm,fill=red!20!white] (char) {#1};}}

\newtheorem{definition}{Definition}[section]

\definecolor{gold}{rgb}{0.99,0.78,0.07}
\usetikzlibrary{arrows,shapes,snakes,automata,backgrounds,positioning,decorations.pathmorphing,
	decorations.markings,calc}

\tikzstyle{dtreenode}=[draw=blue!10!gray,rounded rectangle, minimum size=5mm,fill=blue!10!white]
\tikzstyle{dtreeleaf}=[draw=black!60,minimum width=1cm,minimum height=0.4cm,rectangle,fill=blue!50!white]
\tikzset{every loop/.style={looseness=7}}
\tikzset{
	gluon/.style={decorate,draw=black,
		decoration={coil,amplitude=1pt, segment length=5pt}}
}
\tikzset{
	gluon1/.style={decorate,draw=black,
		decoration={coil,amplitude=3pt, segment length=3pt}}
}
\tikzset{
	gluonew/.style={decorate,draw=black,
		decoration={coil,amplitude=1pt, segment length=2pt}}
}

\tikzset{bicolor/.style args={#1 and #2 and #3}{
		path picture={
			\tikzset{rounded corners=0}
			\fill [#1] (path picture bounding box.south west)
			rectangle
			($(path picture  bounding box.north west)!#3!(path picture bounding
			box.north east)$);
			\fill [#2]
			($(path picture bounding box.south west)!#3!(path picture bounding
			box.south east)$)
			rectangle (path picture bounding box.north east);
}}}

\tikzset{tricolor/.style args={#1 and #2 and #3 and #4 and #5}{
		path picture={
			\tikzset{rounded corners=0}
			\fill [#1] (path picture bounding box.south west)
			rectangle
			($(path picture  bounding box.north west)!#4!(path picture bounding
			box.north east)$);
			\fill [#2]
			($(path picture bounding box.south west)!#4!(path picture bounding
			box.south east)$)
			rectangle
			($(path picture  bounding box.north west)!#5!(path picture bounding
			box.north east)$);
			\fill [#3]
			($(path picture bounding box.south west)!#5!(path picture bounding
			box.south east)$)
			rectangle (path picture bounding box.north east);
}}}

\usepackage{listings}

 \definecolor{dkgreen}{rgb}{0,0.6,0}
 \definecolor{gray}{rgb}{0.5,0.5,0.5}
 \definecolor{mauve}{rgb}{0.58,0,0.82}

 \definecolor{deepblue}{rgb}{0,0,0.5}
\definecolor{deepred}{rgb}{0.6,0,0}
\definecolor{deepgreen}{rgb}{0,0.5,0}

\DeclareFixedFont{\ttm}{T1}{txtt}{m}{n}{9}  
\DeclareFixedFont{\ttb}{T1}{txtt}{bx}{n}{9}

\definecolor{cadmiumgreen}{rgb}{0.0, 0.42, 0.24}
\definecolor{verde}{rgb}{0.25,0.5,0.35}
\definecolor{jpurple}{rgb}{0.5,0,0.35}
\definecolor{darkgreen}{rgb}{0.0, 0.2, 0.13}

\usepackage[ruled,vlined,linesnumbered,lined]{algorithm2e}

\title{Detecting and Understanding Real-World Differential Performance Bugs in Machine Learning Libraries}

\begin{document}

\author{Saeid Tizpaz-Niari}
\affiliation{\institution{University of Colorado Boulder}}
\email{saeid.tizpazniari@colorado.edu}

\author{Pavol {\v C}ern\'y}
\affiliation{\institution{TU Wien}}
\email{pavol.cerny@tuwien.ac.at}

\author{Ashutosh Trivedi}
\affiliation{\institution{University of Colorado Boulder}}
\email{ashutosh.trivedi@colorado.edu}

\begin{abstract}
Programming errors that degrade the performance of systems are widespread, yet
there is very little tool support for finding and diagnosing these bugs. We
present a method and a tool based on {\em differential performance analysis} ---
we find inputs for which the performance varies widely, despite having the same
size. To ensure that the differences in the performance are robust
(i.e. hold also for large inputs), we compare the performance of not only
single inputs, but of classes of inputs, where each class has similar inputs
parameterized by their size. Thus, each class is represented by a
performance function from the input size to performance.
Importantly, we also provide an explanation for why the performance
differs in a form that can be readily used to fix a performance bug.

The two main phases in our method are discovery with fuzzing and
explanation with decision tree classifiers, each of which is
supported by clustering.
First, we propose an evolutionary fuzzing algorithm to generate
inputs that characterize different performance functions.
For this fuzzing task,
the unique challenge is that we not only need the input class with
the worst performance, but rather a set of classes exhibiting differential
performance. We use clustering to merge similar input classes
which significantly improves the efficiency of our fuzzer.
Second, we explain the differential performance in terms of program
inputs and internals (e.g., methods and conditions).
We adapt discriminant learning approaches
with clustering and decision trees to
localize suspicious code regions.

We applied our techniques on a set of micro-benchmarks and real-world machine
learning libraries. On a set of micro-benchmarks, we show that our approach
outperforms state-of-the-art fuzzers in finding inputs to characterize
differential performance. On a set of case-studies, we discover and explain
multiple performance bugs in popular machine learning frameworks, for instance
in implementations of logistic regression in \texttt{scikit-learn}. Four of these
bugs, reported first in this paper, have since been fixed by the developers.
\end{abstract}

\maketitle

\section{Introduction}
\label{sec:intro}
\begin{figure*}[t!]
  \centering
  \begin{minipage}[t]{0.24\textwidth}
  \includegraphics[width=\textwidth]{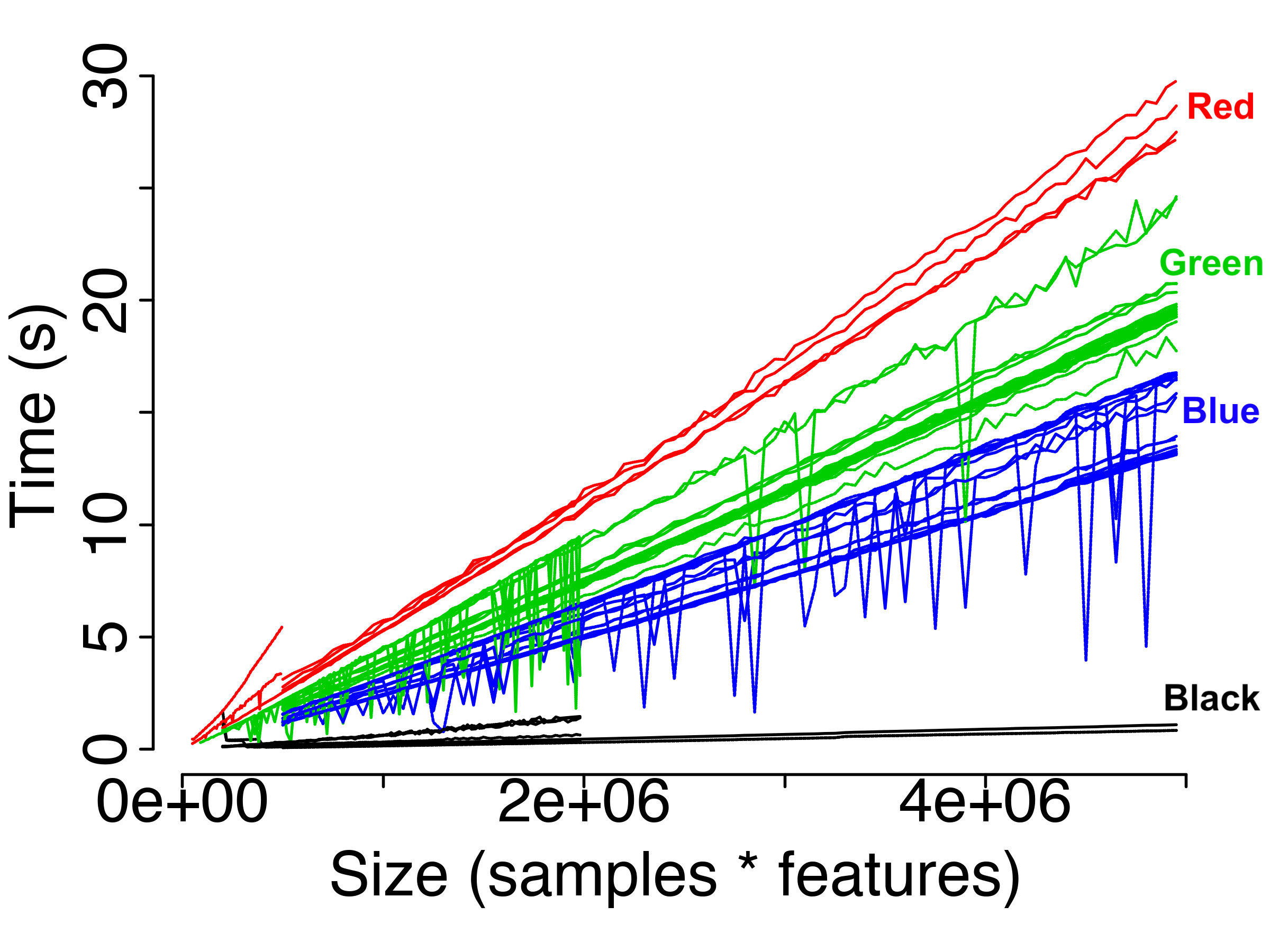}
  \end{minipage}
  \hfill
  \begin{minipage}[t]{0.23\textwidth}
  \scalebox{0.5}{
    \begin{tikzpicture}[align=center,node distance=1cm,->,thick,
        draw = black!60, fill=black!60]
      \centering
      \pgfsetarrowsend{latex}
      \pgfsetlinewidth{0.3ex}
      \pgfpathmoveto{\pgfpointorigin}

      \node[dtreenode,initial above,initial text={}] at (0,0) (l0)  {
        solver};
      \node[dtreeleaf,bicolor={black and black and 0.99}] at (-2,-1.5)
      (l1) {};
       \node[dtreenode] at (0, -1.5) (l2)
        {multi-class};
      \node[dtreenode] at (-1, -3.0) (l3)
       {penalty};
       \node[dtreenode] at (3, -3.0) (l4)
        {penalty};
       \node[dtreeleaf,bicolor={green and black and 0.95}] at (0, -4.5)
       (l6) {};
       \node[dtreenode] at (1.0, -4.5) (l7)
        {C};
       \node[dtreenode] at (3, -4.5) (l8)
        {C};

        \node[dtreeleaf,bicolor={red and green and 0.8}] at (-2, -4.5)
        (l5) {};
        \node[] at (0.0, -6) (l11)
         {};
         \node[dtreeleaf,bicolor={blue and blue and 0.99}] at (1.0, -6)
         (l12) {};
         \node[] at (2.5, -6) (l13)
          {};
        \node[] at (3.0, -6) (l14)
         {};
         \node[] at (3.0, -1.5) (l15)
          {};

        \path[->]  (l0) edge  node [left,pos=0.4] {=`lbfgs'~} (l1);
        \path  (l0) edge  node [right,pos=0.7] {=`saga'} (l2);
        \path  (l0) edge[dotted]  node [right,pos=0.4] {=`sag'or`liblinear'\\or`newton-cg'} (l15);
        \path  (l2) edge  node [right,pos=0.7] {~~=`multi\\nomial'} (l3);
        \path  (l2) edge  node [right,pos=0.7] {~=`ovr'} (l4);
        \path  (l3) edge  node [right] {=`l1'} (l5);
        \path  (l3) edge  node [right] {=`l2'} (l6);
        \path  (l4) edge  node [left] {=`l1'} (l7);
        \path  (l4) edge  node [right] {=`l2'} (l8);
        \path  (l7) edge[dotted]  node [left] {~$>$5.5} (l11);
        \path  (l7) edge  node [right,pos=0.2] {~$\leq$5.5} (l12);
        \path  (l8) edge[dotted]  node [left,pos=0.7] {~$>$5.5} (l13);
        \path  (l8) edge[dotted]  node [right] {~$\leq$5.5} (l14);

    \end{tikzpicture}
  }
  \end{minipage}
  \hfill
  \begin{minipage}[t]{0.23\textwidth}
  \scalebox{0.5}{
    \begin{tikzpicture}[align=center,node distance=1cm,->,thick,
        draw = black!60, fill=black!60]
      \centering
      \pgfsetarrowsend{latex}
      \pgfsetlinewidth{0.3ex}
      \pgfpathmoveto{\pgfpointorigin}
      \node[dtreenode,initial above,initial text={}] at (0,0) (l0)  {
        solver};
      \node[dtreeleaf,bicolor={blue and blue and 0.99}] at (-3.5,-2) (l1) {};
      \node[dtreenode] at (-0.6, -2) (l2)
       {dual};
       \node[dtreenode] at (1, -2) (l3)
        {multi-class};
        \node[dtreeleaf,bicolor={black and black and 0.99}] at (-1.5, -4)
        (l5) {};
        \node[dtreeleaf,bicolor={red and green and 0.7}] at (-3.0, -4)
        (l6) {};
        \node[dtreeleaf,bicolor={black and black and 0.99}] at (3.0, -4)
        (l7) {};
        \node[dtreenode] at (0.5, -4) (l8)
        {penalty};
        \node[dtreenode] at (-0.5, -5.5) (l9)
        {tol};
        \node[dtreeleaf,bicolor={black and blue and 0.85}] at (1.5, -5.5)
        (l10) {};
        \node[dtreeleaf,bicolor={blue and blue and 0.99}] at (-1.5, -7.0)
        (l11) {};
        \node[dtreeleaf,bicolor={black and black and 0.99}] at (0.5, -7.0)
        (l12) {};
        \path[->]  (l0) edge  node [left,pos=0.4] {=`sag'~} (l1);
        \path  (l0) edge  node [left, near end] {=`liblinear'} (l2);
        \path  (l0) edge  node [right, near end] {=`newton-cg'} (l3);
        \path  (l2) edge  node [right] {=False} (l5);
        \path  (l2) edge  node [left] {=True~~} (l6);
        \path  (l3) edge  node [right] {= `ovr'} (l7);
        \path  (l3) edge  node [right,pos=0.7] {=`multi\\nomial'} (l8);
        \path  (l8) edge  node [right] {= `l1'} (l10);
        \path  (l8) edge  node [left] {=`l2'} (l9);
        \path  (l9) edge  node [left] {~$=$ 0.0} (l11);
        \path  (l9) edge  node [right] {$>$ 0.0~} (l12);

    \end{tikzpicture}
  }
  \end{minipage}
  \hfill
  \begin{minipage}[b]{0.23\textwidth}
  \scalebox{0.5}{
    \begin{tikzpicture}[align=center,node distance=1cm,->,thick,
        draw = black!60, fill=black!60]
      \centering
      \pgfsetarrowsend{latex}
      \pgfsetlinewidth{0.3ex}
      \pgfpathmoveto{\pgfpointorigin}

      \node[dtreenode,initial above,initial text={}] at (0,0) (l0)
      {optimize.if np.max(absgrad)};
      \node[dtreeleaf,bicolor={blue and blue and 0.99}] at (-1.0, -1.5)
      (l1) {};
      \node[dtreeleaf,bicolor={black and blue and 0.85}] at (1.0, -1.5)
      (l2) {};

      \path[->]  (l0) edge  node [left,pos=0.4] {$ > 14$} (l1);
      \path  (l0) edge  node [right, pos=0.4] {$\leq 14$} (l2);
    \end{tikzpicture}
  }
  \end{minipage}
  \caption{
    From left to right: (a) four classes of performances discovered by fuzzing
    and clustering, (b) decision tree learned in the space
    of parameter features (part 1), (c) decision tree learned in the space
    of parameter features (part 2), (d) decision tree learned using the internal
    features to explain differences in performance of `newton-cg' solver.}
    \label{fig:logistic-regression-clustered}
    \label{fig:logistic-regression-classifier-inp}
    \label{fig:logistic-regression-classifier-internal}
\end{figure*}
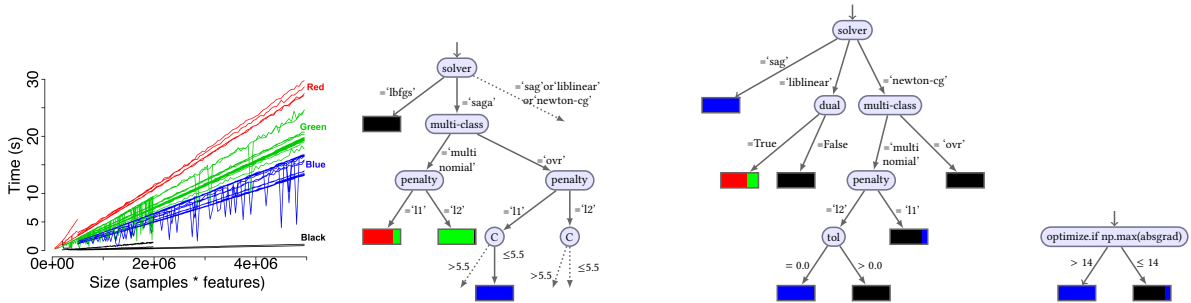

The defects in software developments can lead to severe
performance degradations and waste valuable system resources
such as CPU cycles.
Moreover, studies have shown that such performance bugs are widespread
in real-world applications~\cite{aaai18,song2014statistical,olivo2015static}.
How can a user of a program recognize a performance bug? Most often, they can suspect a bug if the program has different performance for similar inputs.
Recently, such bugs have been reported by users of machine learning libraries~
\cite{rnd-forest-issue,logistic-regressio-issue,pair-wise-distance-issue,train-slow-cpu-issue}.
For example, a user of random forest regression reported that
the trees with `mae' criterion are slower than those with `mse'. Once such an
issue is discovered, how can a maintainer of a library know whether the difference is  the result of a bug, or if it is inherent in the problem being solved?

We present a method and a tool to address the challenges we just described.
We use \textit{fuzzing} to generate inputs that uncover performance
bugs and \textit{discriminant analysis} to explain the differences in
performance. The fuzzing part of our study generalizes evolutionary-based fuzzing
algorithms~\cite{AFL} in two ways. {\em First, we consider multiple populations
of inputs.}
Each population corresponds to a {\em simple path}
\footnote{A path is simple if it contains an edge at most once.} in the
program's control-flow graph. An
execution that takes a loop $5$ times is represented by the same simple path as
an execution that takes the loop $7$ times. Therefore, each class (or population)
of inputs is represented by a function from the input size to performance.
{\em Second, rather than searching for the single worst-case input class, our fuzzer
explores classes of inputs with significant performance differences.} To do so,
we propose an evolutionary algorithm such that it both models
the performance as a function of input size and finds a set of classes
of inputs with diverse performance efficiently.
The key idea for the efficiency is to combine evolutionary fuzzing algorithms with
the functional data {\em clustering}~\cite{jacques2014functional}.
The clustering is used to focus the search to retain representative
paths from a few prominent clusters of paths
instead of repeatedly exploring paths with similar performance.

Once a suspicious performance abnormality has been uncovered,
the next step is to pinpoint the root causes for the differential
performance. For this task, we adapt techniques from discriminant learning
algorithms~\cite{aaai18,breiman1984classification} to
functional data~\cite{ramsay2006functional}. The
causes of the diverse performance are explained using features from the space of
program (hyper)parameters and from the space of program internals  (such as the
number of invocations of a particular method). We learn a discriminant model
that shows what features are the same inside a cluster and what features
distinguish one from another.

Previous works in performance fuzzing consider the worst-case algorithmic
complexity~\cite{petsios2017slowfuzz,lemieux2018perffuzz} and
search for a single input with the significant resource usage. Differential fuzzing
is used in security to detect timing side channels in Java
applications~\cite{DBLP:conf/icse/nilizadeh,FuncSideChan18}.
To the best of our knowledge, this is the first
work to automatically generate inputs to characterize multiple performance
classes. Also, this work utilizes the inputs from the fuzzer to
automatically find code regions contributing to the differential performance,
while previous works focusing on performance bug localization assume that
the interesting inputs are given~\cite{aaai18,song2014statistical}.

We apply our approach on a set of micro-benchmark
programs and larger machine learning libraries.
On a set of micro-benchmarks that include well-known sorting,
searching, tree, and graph algorithms~\cite{algorithm-book},
we demonstrate that although our approach is slower in generating inputs,
it outperforms other fuzzing techniques~\cite{lemieux2018perffuzz,petsios2017slowfuzz}
in characterizing differential performance.
On a set of eight larger machine learning tools and libraries, we find multiple
previously-unreported performance bugs in widely used libraries such
as in the implementation of logistic regression in scikit-learn
framework~\cite{scikit-learn}. We have reported these bugs, and four of them have
since been fixed by the developers.

\noindent The key contributions of our paper are:
\begin{compactitem}

\item We extend fuzzing algorithms to functional data and, crucially, use
clustering {\em during} the fuzzing process to efficiently find classes of inputs
with widely different performance. We show the importance of clustering
during the fuzzing phase through comparison to state-of-the-art fuzzers.

\item We use the discriminant learning approach alongside the fuzzing
to find the root cause of performance issues.

\item We implement our approach in the tool \toolname and evaluate it on eight
machine learning libraries. We show the usefulness of \toolname in finding and
explaining multiple performance bugs such as in
scikit-learn libraries~\cite{scikit-learn}.

\end{compactitem}

\section{Overview}
\label{sec:overview}
First, we show how \toolname can be used to detect a performance
bug in a popular machine learning library. Then, we describe the components
of \toolname.

\noindent \textbf{A) Applying \toolname on Logistic Regression.}
Logistic regression in scikit-learn~\cite{scikit-learn} is a popular classification model that supports
various solvers and penalty functions. We refer
the reader to \cite{logistic-regression-explanation}
for more information about the functionality of this classifier.
We analyzed the performance of logistic
regression\footnote{\url{https://scikit-learn.org/stable/modules/generated/sklearn.linear_model.LogisticRegression.html}}.

\noindent \textbf{Task.} We apply \toolname to automatically
generate inputs that find diverse classes of
performances. If there are multiple classes (more than one cluster),
we explain the performance issues with \toolname.

\noindent \textbf{Generating interesting inputs.}
We run the fuzzer for about two hours, and it generates $185$ sets of inputs, where each
input corresponds to a unique path in the control flow graph (CFG) of the
program.  Since each set includes multiple inputs, the performance of each path
gives rise to functions. The fuzzer thus provides $185$ different performance
functions. The coverage of \toolname is $78$\% where it covers $2.1$K LoC from
almost $2.7$K LoC.

\noindent \textbf{Clustering performance functions.}
Given the performance functions, we
use functional data clustering~\cite{jacques2014functional}
algorithms to group them into a smaller number of clusters.
Figure~\ref{fig:logistic-regression-clustered} (a) shows
that the performance functions are clustered into 4 groups.

\begin{figure*}[t!]
  \includegraphics[width=\textwidth]{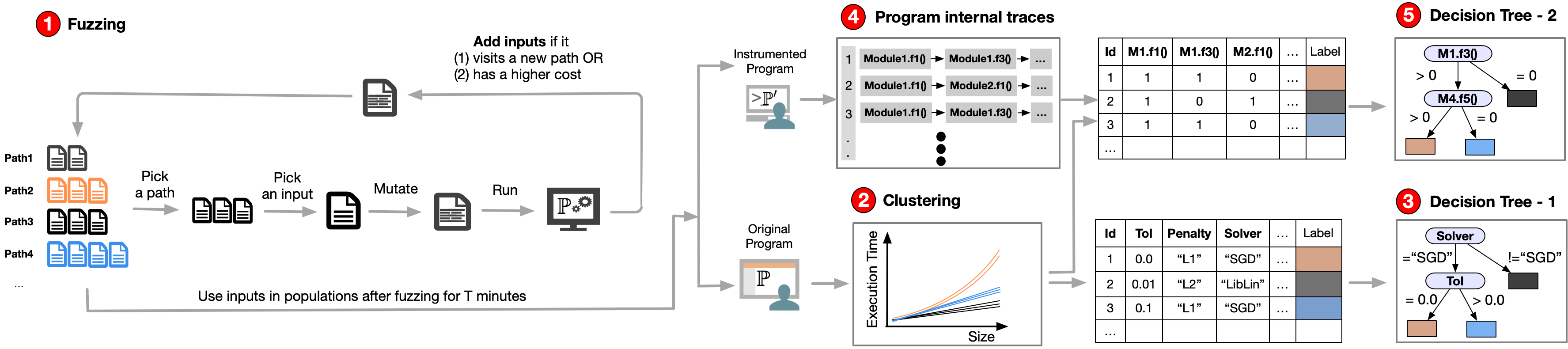}
  \caption{The workflow of \toolname. The three main tasks are (1) fuzzing to
  generate interesting inputs, (2) clustering to find classes of performance
  functions, and (3) explaining different performance classes with decision tree-1
  in the space of program inputs and decision tree-2 in the space of program
  internals.}
  \label{fig:DPFUZZ-overview}
\end{figure*}

\noindent \textbf{Explaining the clusters of performances.}
Given the set of inputs and their performance labels,
we use the discriminant learning approach~\cite{tizpaz2017discriminating}
with decision tree algorithms to explain the
differences between performance clusters in terms of program hyper-parameters
and program internal features.

The decision trees in Figure~\ref{fig:logistic-regression-classifier-inp}
(b) and (c) show the discriminant models in the space of
hyper-parameters.
One particular (expected) observation is that different solvers have different
performance. However, more interesting parts
are those with the differential performance in the same solver.
For example, Figure~\ref{fig:logistic-regression-classifier-inp} (c) shows the
inputs with `newton-cg' solver can be in black (fast) or blue (slow) clusters.
In particular,
there is an unexpected performance differences between `tol' = 0.0
and `tol' $>$ 0.0 (even for `tol'  = 0.000001). The inputs with `tol' = 0.0
follow (slow) blue cluster whereas the inputs with positive
values follow (fast) black cluster.

The next step is to help the library maintainer explain the differences between
the fast and slow clusters in terms of library internals such as function calls.
\toolname obtains 1,673 features about the internals of logistic regression
through instrumentations. The decision tree model in
Figure~\ref{fig:logistic-regression-classifier-inp} (d) shows the
discriminant model for `newton-cg' solver in the space of the internal features
(chosen from a set of $5$ accurate models).
The model shows that the number of calls to `\texttt{if np.max(absgrad)}' in
the \texttt{optimize} module is the discriminant that distinguishes the
blue and black clusters:

\begin{scriptsize}
\begin{lstlisting}
  while k < maxiter:
    # call to compute gradients
    absgrad = np.abs(fgrad)
    if np.max(absgrad) < tol:
      break
    ...
    # inner loop: call to compute loss
\end{lstlisting}
\end{scriptsize}
The outer loop of Newton iterations will be unnecessary taken
even if \texttt{np.max(absgrad)} becomes 0.0 for the case where
the tolerance (`tol') is set to zero. This wastes CPU resources
by calling to compute the gradient, Hessian, and loss unnecessarily.
We mark this as performance bug \circled{1}. With this error localization,
the library maintainer can see that the fix is to replace the strict
inequality with a non-strict one. We reported this bug~\cite{logistic-regression-bug},
and the developers have confirmed and fixed it
using this information~\cite{logistic-regression-fix}.

Another interesting performance issue in logistic regression
is related to `saga' solver. \toolname automatically found the issues in
the solver that was already reported in the issue
database~\cite{logistic-regressio-issue}.
Figure~\ref{fig:logistic-regression-classifier-inp} (b) shows the discriminant
learned for `saga' solver.

\noindent \textbf{Comparison with the existing fuzzers.}
The existing performance fuzzers such as SlowFuzz~\cite{petsios2017slowfuzz}
and PerfFuzz~\cite{lemieux2018perffuzz} are looking for the worst-case execution
time, and it makes them unlikely to find differential performance bugs.
In Figure~\ref{fig:logistic-regression-classifier-inp} (a), the worst-case behavior
(the red function) is not related to the bug found by DPFuzz.
The performance bug \circled{1} is found because of the differences between blue
and black functions, and neither of them is the worst-case behavior.
Exploring and explaining various classes of performance are the novelty
in \toolname that find these subtle performance bugs.

\noindent \textbf{B) Inside \toolname}
Figure~\ref{fig:DPFUZZ-overview} shows different components of \toolname.
The \textit{first} component of \toolname is to generate inputs. For this part,
we extend the evolutionary fuzzing algorithms~\cite{AFL,libFuzzer}
with functional data analysis and clustering~\cite{ramsay2006functional,jacques2014functional}.
Our fuzzing approach considers multiple populations
(one population per a distinct path in the CFG).
Then, it picks a cluster, a path from the cluster, and
an input from the selected path. Next, it mutates and crossovers the
input and runs the input on the target program. This returns the cost of
executions (either in terms of actual execution times or in terms of executed
lines), and the path characterization. The fuzzing approach adds a new
input to the populations if the input has visited a new path in the program or
the input has achieved higher costs in comparison to the inputs
in the same simple path. The fuzzing stops after $T$ time units and provides
generated inputs for debugging.

The \textit{second} component of \toolname is to characterize
different performance classes.
The set of inputs in a path (or a population) defines
a performance function varied in the input size.
Given $n$ paths (corresponds to $n$ performance functions),
\toolname applies clustering algorithms to partition theses functions into $k$ classes of
performances ($k \leq n$). The clustering is primarily based on the non-parametric functional
data clustering~\cite{jacques2014functional} with $l_1$ distance.
The clustering finds similar input classes and separates classes
with significant performance differences.
The plot in Figure~\ref{fig:logistic-regression-clustered} (a) is generated
as a result of fuzzing and clustering steps.

The \textit{third} component of \toolname is to explain the differential
performance in terms of program inputs and internals. The CART decision tree
inference~\cite{breiman1984classification} is mainly used to obtain the
explanation models. In the space of program inputs,
the features are input parameters such as the value of ``solver'', and
the labels are the performance classes from the clustering algorithm. The
decision tree-1 in Figure~\ref{fig:DPFUZZ-overview} is a sample model in
the space of program inputs. The model shows
what input parameters are common in the same cluster and what the parameters
distinguish different clusters. The models in
Figure~\ref{fig:logistic-regression-clustered} (b) and (c) are produced from
this step. Using this decision tree, the user may realize that
all or some aspects of differential performance are unexpected.
The idea is to find code regions that contribute to the
creation of an unexpected performance.

In the space of program internals,
the instrumentation of target programs is used to
obtain program internal traces.
For this aim, tracing techniques~\cite{Trace}
are applied to generate a trace of execution for inputs from either the whole
population or relevant to the unexpected performance.
Next, we gather program internal features from these traces. The
features used in this work are the number of calls to functions, conditions,
and loops. Given these program internal features and the performance class
labels (from the clustering algorithm), the problem of localizing code
regions (related to the differential performance) becomes a standard
classification problem.
The decision tree-2 in Figure~\ref{fig:DPFUZZ-overview} is learned in the
space of program internal features that show what properties of program
internals are most likely responsible for differential performance.
Figure~\ref{fig:logistic-regression-clustered} (d)
is produced from this step.

\section{Problem Statement}
\label{sec:definition}
Following the work of Hartmanis and Stearns~\cite{HS65},
it is customary to
characterize the resource complexity of a program as a function of the
program input size characterizing the worst/average/best performance.
However, often there are latent modes in the program inputs characterizing
widely different complexity classes, and knowing the existence and explanation
of their differences will serve as a debugging aid for the developers
and users.
We study the problem of discovery (via evolutionary fuzzing) and
explanation (via classifications) of latent resource
complexity classes.
To formalize our problem, we use the following abstract model:
\begin{definition}[Performance Abstraction of a Program]
  An abstract {\it performance  model} $\sPp$ of a program $\Pp$ is a tuple
  $\sPp = (X, Y, \Oo, \pi)$ where:
  \begin{itemize}
  \item
    $X {=} \set{x_1, \ldots, x_n}$ is the set of
    {\it input} variables characterizing the input space $\Dd_X$ of the program,
  \item
    $Y = \set{y_1, y_2, \ldots, y_m}$ is the set of {\it trace} variables
    characterizing the execution space $\Dd_Y$ of the program,
  \item
    $\Oo: \Dd_X \to \Dd_Y$ is the {\it functional output} of the program
    summarizing effect of the input on trace variables, and
  \item
    $\pi: \Dd_X \to \Rplus$ is the performance (running time or memory usage)
    of the program.
  \end{itemize}
\end{definition}

Following the asymptotic resource complexity convention, we assume the existence
of a function $|\cdot| : \Dd_X \to \Nat$ providing an estimate of the input size. For ML applications, the size can be a product of number of samples and
features.
For a subset $S \subseteq \Dd_X$ of the input space, we define its performance
class as the function $\pi_S: \Nat \to \Rplus$ defined as worst-case complexity
given below:
\[
\pi_S \rmdef n \mapsto \sup_{\vx \in S \::\: |\vx| = n} \pi(\vx).
\vspace{-0.2em}
\]
Note that $\pi_S$ is a partial function for finite sets $S \subset \Dd_X$.
Let $\Pi$ be the set of all performance classes.
We introduce the distance function
$\vd{p}: \Pi \times \Pi \to \Rplus$ on the set of performance classes
parameterized with $p$. The typical choice for $d_p$ is $p$-norm ($p = 1, 2, \infty$)
and $R^2$ coefficient of determinations.

\begin{definition}[Differential Performance Fuzzing]\label{def:problem-1}
Given a program $\Pp = (X, Y, \Oo, \pi)$ and a separation bound $\varsigma > 0$,
the {\it differential performance fuzzing problem} is to find a set partition
$P = \set{D_1, D_2, \ldots, D_k}$ of $\Dd_X$ such that for every $A, B \in P$
we have that $d_p(\pi_{A}, \pi_B) > \varsigma$.
\end{definition}

Even when the input space is finite, the differential performance fuzzing
problem requires an exhaustive search in the exponential set of subsets of input
space, and hence is clearly intractable.
In Section~\ref{sec:diff-perf-fuzz}, we present an evolutionary algorithm
to solve the problem by restricting
the performance classes to polynomial functions.

Once the fuzzer reports a partition of the input space into sets with
distinguishable performance classes, the debugging problem is to find an
explanation of the distinction between various performance classes.
Oftentimes, this explanation can not be reported based solely on the values of
the input variables and is a function of syntactic structure ($Y$ variables) of the program.
To enable such explanations, we study the discriminant learning problem, where
the goal is to learn the differences between various partitions of the inputs as
predicates over the trace variables.

A predicate over the execution space $\Dd_Y$ is a function $\phi : \Dd_Y \to
\set{\texttt{T},\texttt{F}}$.
Let $\Phi(\Dd_Y)$ be the set of predicates over $\Dd_Y$.
Given an input partition $P = \set{D_1, D_2, \ldots, D_k}$ of a program
$\sPp = (X, Y, \Oo, \pi)$, a discriminant is function $\Delta: P \to \Phi$ such that
for every $D_i \in P$ we have that $\vx \in D_i$ implies $\Delta(D_i)(\Oo(\vx)) =
\texttt{T}$.

\begin{definition}[Differential Performance Debugging]\label{def:problem-2}
Given a program $\Pp = (X, Y, \Oo, \pi)$ and an input partition $P =
\set{D_1, \ldots, D_k}$, the {\it differential performance
  debugging problem} is to find a discriminant $\Delta: P {\to}
\set{\texttt{T},\texttt{F}}$ of $P$.
\end{definition}

Tizpaz-Niari et al.~\cite{tizpaz2017discriminating} showed that the differential
performance debugging problem is already NP-hard for programs with a finite set
of inputs. In Section~\ref{sec:diff-perf-debug}, we present a data-driven approach to
learn discriminants as decision trees.

\section{Data-Driven Approach}
\label{sec:approach}
To address problems~\ref{def:problem-1}~and~\ref{def:problem-2},
we propose a data-driven
approach in two steps.
First, we extend gray-box evolutionary
fuzzing algorithms~\cite{AFL,libFuzzer} to generate performance
differentiating inputs.
Our fuzzer is equipped with clustering to identify
performance classes while generating inputs.
Second, we use discriminant learning~\cite{aaai18} to
pinpoint code regions connected to the differential performance.

\subsection{Differential Performance Fuzzing}
\label{sec:diff-perf-fuzz}
The overall fuzzing algorithm is shown in Algorithm~\ref{alg:overall}.
Given a program $\Pp$, the goal is to discover widely varying
performance classes.
Next, we describe important components of our evolutionary
search based Algorithm~\ref{alg:overall}.

\begin{algorithm}[t!]
{
	\DontPrintSemicolon
	\KwIn{Program $\Pp$, initial seed $X$,
		max. number of iterations $N$,
		steps to do clustering $M$,
		the tolerance $\epsilon$,
		the separation $\varsigma$,
		num. of clusters $k$.
	}
	\KwOut{Inputs/clusters manifest performance of $\Pp$.}

	\For{$x \in X$}
	{
		$path, cost$ $\gets$ \texttt{run}($\Pp$, $x$)

		$Cov[path]$.\texttt{add}($cost$), $Pop[path]$.\texttt{add}($s$)
	}

	$clusters$ $\gets$ \texttt{clust}($Cov,Pop,\epsilon$)

	$step$ $\gets$ 1

	\While{step ${\leq N} \land$ $|clusters_\varsigma| < k$}{

		\If{$step~\%~M$ = 0}{
			$clusters$ $\gets$ \texttt{clust}($Cov,Pop,\epsilon$)
		}

		$cluster$ $\gets$ \texttt{choice$_{R}$}($clusters$)

		$path$ $\gets$ \texttt{choice$_{W}$}($cluster$)

		$x$ $\gets$ \texttt{choice$_{W}$}($Pop[path]$)

		$x'$ $\gets$ \texttt{mutate}($x$)

		$path', cost'$ $\gets$ \texttt{run}($\Pp$, $x'$)

		\If{$path' \not \in Cov.keys()$}{
				$Cov[path']$.\texttt{add}($cost'$),$Pop[path']$.\texttt{add}($inp'$)
		}
		\ElseIf{\texttt{promisingInput}($cost',path',n$)}{
				$Cov[path']$.\texttt{add}($cost'$),$Pop[path']$.\texttt{add}($inp'$)
		}

		$step$ $\gets$ $step$ + 1
	}

	\Return $Cov, Pop, clusters$.

	\caption{\textsc{\toolname: Evolutionary fuzzing for differential performance.}}
	\label{alg:overall}
}
\end{algorithm}

\noindent\textbf{Cost Measure}.
\label{subsec:fitness}
The performance (cost) model $\pi$ summarizes
the resource usages. We consider both 1) abstractions of resource
usages such as the number of lines executed and 2) concrete
resource usages such as the execution times.

\noindent\textbf{Trace Summary}.
We consider an instantiation of trace summary function $\Oo$ for fuzzing where
the function $\Oo$ takes an input ${x \in X}$ and returns a set
of edges in the control flow graph (CFG) visited during
the execution of the input $x$.

\noindent\textbf{Path Model}.
We characterize the subset $S$ of input space using information from program
traces. In particular, we use the path information to determine if
two inputs are in the same class ${s \in S}$.
We represent paths using the hash values of their ids.
For each edge in the CFG,
we consider a unique edge id. Then, we apply a hash function
$H$ on the set of edge ids
visited for executing an input. Two distinct inputs ${x_1,x_2 \in X}$ are in the same
performance class if ${H(y_1)=H(y_2)}$
where ${y_1=\Oo(x_1)}$ and ${y_2=\Oo(x_2)}$.

\noindent\textbf{New Input}.
We add an input to the population if the path is new, i.e.
the id of the path is not in the coverage key
(line 14 in Algorithm~\ref{alg:overall}), or the input
has a higher cost in comparison to other inputs in the same path
(line 16 in Algorithm~\ref{alg:overall}).

\noindent\textbf{Functional Data Clustering}.
We use non-parametric functional data clustering
algorithms~\cite{jacques2014functional} to cluster paths into a few
similar performance groups (line 4 and 8 in Algorithm~\ref{alg:overall}).
For a set of inputs $x_1,\ldots,x_n$ mapped to
the same path $h$, i.e. ${H(y_1)=\ldots=H(y_n)=h}$,
we fit linear ${a|x|+b}$ and polynomial ${a|x|^b}$ functions~\cite{goldsmith2007measuring}
to model the performance function $\pi_h$.
Then, we calculate the $l_{1}$ distances between
the performance functions and apply \texttt{KMeans}
clustering~\cite{lloyd1982least}
with the tolerance bound ${\epsilon > 0}$
on the distance matrix to
partition the paths into $k$ clusters $\Sigma$.
The clustering guarantees the following condition:
if two paths (and their corresponding
performance functions) are in the same cluster (${h,h' \in \Sigma_i}$), then
$d_1(\pi_{h},\pi_{h'}) \leq \epsilon$.

The \texttt{clust} function in Algorithm~\ref{alg:overall}
works as the following: starting with one
cluster (${k=1}$), we apply KMeans clustering and check
if the condition holds, i.e. all the functions in the
same cluster are $\epsilon$ close to each other. If this is the
case, we return the clusters.
Otherwise, we increase $k$ to $k+1$ and run the algorithm again.
The clustering algorithm helps explore (few) paths with distinguishable
performance functions as opposed to (too many) paths in the program.
Each cluster contains one or
more paths that have similar performance behaviors. The selection
function (line 10 in Algorithm~\ref{alg:overall}) chooses a path inside
a cluster based on weighted probabilities where a path with higher score
of cost has a better chance of selection. Similar criterion has used to
choose an input from the set of possible inputs inside the path population
(line 11 in Algorithm~\ref{alg:overall}).

\noindent\textbf{Mutations and Crossover.}
\label{subsec:mutate}
We consider 8 well-established~\cite{AFL,petsios2017slowfuzz,lemieux2018perffuzz}
mutation operations
to guide the search algorithm.
We also consider crossover operation where it mixes the current
input with another input from the population.

\noindent\textbf{Termination Conditions}.
The Algorithm~\ref{alg:overall} terminates either if it reaches
to the maximum steps $N$ or there are $k$ clusters with
significant performance differences
(line 6 in Algorithm~\ref{alg:overall}).

\subsection{Differential Performance Debugging}
\label{sec:diff-perf-debug}
Given the set of inputs $Pop{=}{\set{{X_1},{\ldots},{X_k}}}$
and their performance label $\Sigma$
from the fuzzing step, the algorithm~\ref{alg:overall-2}
explains the differential performance in the inputs.
First, the user of \toolname
(optionally) runs the clustering algorithm with the parameter
$k$ to obtain performance clusters
(line 1 in Algorithm~\ref{alg:overall-2}).
The debugging procedure uses the inputs $Pop$ as features and their clusters
as labels to learn a decision tree model that explains
the differential performance in the space of input parameters.
We use CART decision tree algorithms~\cite{breiman1984classification} to learn
the set of predicates $P$ in the space of input parameters
(line 3 in Algorithm~\ref{alg:overall-2}).
For example, in the decision tree of Figure~\ref{fig:logistic-regression-classifier-inp}, \texttt{solver}=`saga'${\land}$\texttt{multi-class}=`multinomial'
${\land}$ \texttt{penalty}=`l2' is the predicate for the green cluster
such that inputs satisfying these
parameters belong to the green cluster.

The critical step in debugging is to explain the differences based on the program
internals. For this step, the inputs $Pop$ are feed into the instrumented program
$\Pp'$ that generates program internal features such as whether a method or
a condition are invoked and how many times they are called
(line 4 in Algorithm~\ref{alg:overall-2}).
Given the set of (internal) features
$\set{\Pp'(X_1),\ldots,\Pp'(X_k)}$ and their cluster labels $\Sigma$,
the problem of discriminant learning becomes a standard classification problem.
We use CART algorithms to learn
the set of predicates $\Phi$ in the space of program internal features
(line 5 in Algorithm~\ref{alg:overall-2}).
These predicates partition the space of internal features
into hyper-rectangular sub-spaces $\set{\phi_1,\ldots,\phi_k}$
such that if ${H(\Oo(x)) \in \Sigma_j}$, then $\Pp'(x){\models}\phi_j$,
i.e., the predicate $\phi_j$ evaluates to true for the evaluation
of input $x$ trace feature given that the input $x$ is in the performance cluster
$j$. For example, a predicate that evaluates whether a
particular method is invoked determines
the complexity of its performance class.
In Figure~\ref{fig:logistic-regression-classifier-inp} (d),
the predicate based on whether the number of calls to
the condition \texttt{np.max(absgrad)} is more than 14 distinguishes the blue
and black clusters. The debugger uses this information to localize regions
in the code and to potentially fix a performance bug.

\begin{algorithm}[t!]
{
	\DontPrintSemicolon
	\KwIn{Program $\Pp$, instrumented program $\Pp'$, desired clusters $k$,
	inputs $Pop$, the class $label$.
	}
	\KwOut{The set of predicates $P,\Phi$ explain differential performances.}

	$label$ $\gets$ \texttt{clust}($Pop,k$)

	$P$ $\gets$ \texttt{DecisionTree}($Pop,label$)

	$Y$ $\gets$ $\Pp'$($Pop$)

	$\Phi$ $\gets$ \texttt{DecisionTree}($Y,label$)

	\Return $P,\Phi$.

	\caption{\textsc{\toolname: Explaining differential performances.}}
	\label{alg:overall-2}
}
\end{algorithm}

\section{Experiments}
\label{sec:experiment}
\begin{table*}[tbp!]
    \caption{Micro-benchmark results. Comparing \toolname vs SlowFuzz~\cite{petsios2017slowfuzz}
      and PerfFuzz~\cite{lemieux2018perffuzz}.
      Legend: \textbf{\#L}: lines of code, \textbf{\#N}: number of generated samples,
      \textbf{T}:\ fuzzing time (min),
      \textbf{W}: worst-case computation cost (in term of executed lines),
      \textbf{\#P}: number of unique paths,
      \textbf{\#M}: Number of distinct performance functions,
      \textbf{\#K}: Number of functional clusters.
      Note: $M=10^6$ and $K=10^3$.
    }
    \label{tab:fuzzing-variants}
    \resizebox{\textwidth}{!}{
      \begin{tabular}{ | l | r  r | r  r  r  r  r | r  r  r  r  r | r  r  r  r  r|}
      \hline
      &       &    & \multicolumn{5}{c|}{SlowFuzz~\cite{petsios2017slowfuzz}} & \multicolumn{5}{c|}{PerfFuzz~\cite{lemieux2018perffuzz}} &  \multicolumn{5}{c|}{\toolname} \\
      \cline{4-18}
      Algorithm& \textbf{\#L} & \textbf{T}
      & \textbf{\#N} & \textbf{W}   & \textbf{\#P}& \textbf{\#M} & \textbf{\#K}
			& \textbf{\#N} & \textbf{W}   & \textbf{\#P}& \textbf{\#M} & \textbf{\#K}
			& \textbf{\#N} & \textbf{W}   & \textbf{\#P}& \textbf{\#M} & \textbf{\#K} \\ \hline
     	Quick Sort & 80 & 90 & 13.6M & 721 & 5 & 4 & 2 & 10.8M & 716 & 14 & 8 & 3 & 7.8M & 721 & 14 & 11 & 3
       \\ \hline
			3-Ways Q-Sort & 83 & 90 & 12.0M & 756 & 4 & 2 & 1 & 12.1M & 801 & 10 & 6 & 3 & 7.1M & 847 & 10 & 8 & 5 \\ \hline
			InsertionX Sort & 42 & 90 & 15.0M & 496 & 3 & 2 & 1 & 15.2M & 490 & 10 & 4 & 2 & 10.9M & 497 & 10 & 9 & 4 \\ \hline
			Merge Sort & 53 & 90 & 24.9M & 516 & 4 & 1 & 1 & 22.5M & 516 & 6 & 5 & 1 & 10.4M & 516 & 6& 5 & 1 \\ \hline
			Binary Search & 75 & 90 & 11.2M & 530 & 7 & 4 & 1 & 11.2M & 527 & 35 & 21 & 6 & 5.7M & 529 & 34 & 26 & 6  \\ \hline
			Seq. Search & 26 & 90 & 43.8M & 60 & 2 & 2 & 1 & 50.0M & 60 & 6 & 3 & 1  & 13.2M  & 72 & 6 & 4 & 1 \\ \hline
			Boyer Moore & 88 & 90 & 29.2M & 204 & 4 & 0 & 0 & 34.7M & 372 & 8 & 1 & 1 & 9.6M & 372 & 8 & 1 & 1 \\ \hline
			BST Insert & 47 & 90 & 21.2M & 501 & 6 & 3 & 1 & 18.1M & 561 & 13 & 12 & 3 & 7.7M & 566 & 13& 12 & 5 \\ \hline
			Is BST & 141 & 90 & 22.3M & 280 & 14 & 7 & 1 & 26.3M & 224 & 46 & 19 & 2 & 10.0M & 280 & 40 & 25 & 2 \\ \hline
			Prim's MST & 294 & 90 & 7.9M & 1,006 & 10 & 5 & 2 & 7.9M & 975 & 119 & 23 & 5 & 5.3M & 1,020 & 118 & 70 & 11 \\ \hline
    \end{tabular}
 }
\end{table*}

\subsection{Implementation Details}
\noindent \textbf{Environment Setup.}
We use a super-computing machine for running our fuzzer. The machine has a
Linux Red Hat 7 OS with 24 cores of 2.5 GHz CPU each with 4.8 GB RAM.
Since the performance measure for machine learning libraries is the
actual execution times (noisy observations), we re-run the generated inputs
from the fuzzer on a more precise but less powerful NUC5i5RYH machine and
use the measurements of this machine for clustering.
We consider the version 2.7 of python and 0.20.3 of scikit-learn.

\noindent \textbf{Fuzzing and Clustering.}
We implement the fuzzing of \toolname in python by extending the Fuzzing Book
framework~\cite{fuzzingbook2019:index}. The implementation
is over $900$ lines of code and can fuzz both python and Java
applications. The performance measure is the actual running times in
case-studies and the number of executed lines in micro-benchmarks.
We use trace library~\cite{Trace} (python) and Javassist~\cite{chiba1998javassist}
(Java) to model paths and measure performances.
We use numpy polynomial module~\cite{numpy}
to fit performance functions. We implement the KMeans clustering
algorithm using scikit-learn~\cite{scikit-learn}.

\noindent \textbf{Debugging and Instrumentation.}
We instrument python libraries with tracing~\cite{Trace} and
Java applications with Javassist~\cite{chiba1998javassist} to
extract internal features.
We implement the decision tree classifier using CART
algorithm in scikit-learn~\cite{scikit-learn}.

\subsection{Micro-benchmark Results}
We compare our fuzzing technique \toolname
against state-of-the-art performance fuzzers.
For the benchmark, we consider standard sorting, searching, tree, and graph
algorithms from~\cite{algorithm-book}. These benchmarks are
standard programs used to evaluate performance fuzzers.
We consider SlowFuzz~\cite{petsios2017slowfuzz}
and PerfFuzz~\cite{lemieux2018perffuzz} from the literature.
All three fuzzers share the same functionality such as
mutations. The differences are in the population model,
adding a new input to the population,
and choosing an input from the population.

\noindent\textbf{SlowFuzz.} The SlowFuzz~\cite{petsios2017slowfuzz}
aims to find the worst-case
algorithmic complexity. The fuzzing approach has a global population
where it adds a new input to the population if it
achieves a higher cost (performance measure) than any other inputs in the population.
The fuzzing approach chooses an input for mutations from the current
population randomly.

\noindent\textbf{PerfFuzz.} This fuzzing~\cite{lemieux2018perffuzz}
aims to find the worst-case
algorithmic complexity for each entity (such as an edge) in the CFG.
The fuzzing has a global population, and it adds a new
input to the population if the input has visited a new
edge (discovered a new path)
or the input has achieved the highest cost in visiting
at least one edge. It picks an input for mutation based on
whether the
input has had the highest cost for at least one entity.

\noindent\textbf{\toolname.} The fuzzing follows
Algorithm~\ref{alg:overall} where there are multiple populations,
one for each unique path in the CFG. An input is added to the
population if the path induced from it has visited a new edge
in the CFG (forms a new population) or the input has
the highest cost in the
population of this path. \toolname performs clustering
after many steps (set to 1,000 in experiments) and uses the clustering
information to pick an input from the population.

\noindent\textbf{Empirical Research Question.}
For a given program over a fixed time of fuzzing,
we compare the fuzzing techniques on $5$ criteria: the number of generated
inputs, the worst-case computational complexity in terms of executed lines,
the number of visited unique paths, the number
of unique performance functions, and the number of clusters of
performance functions. We repeat each experiment for different
fuzzers $5$ times and report the best results obtained by the fuzzers.
Table~\ref{tab:fuzzing-variants} shows the outcome of each fuzzing
technique on $5$ different criteria for $10$ different algorithms.

\noindent\textbf{Number of generated samples}.
We fix the duration of fuzzing to be 90 minutes for different fuzzers
in all benchmarks. We compare the number of inputs generated
with different fuzzers. Note that the quality of inputs such as
the ones with higher costs of executions affects the number of
generated inputs. In general, SlowFuzz and PerfFuzz could
generate inputs faster in comparison to \toolname. Examples
such as Quick sort, Merge sort, and Binary search provide
fair comparison in terms of generated inputs since all fuzzers
have similar performances in finding worst-case execution times.
The slowdown in \toolname is almost
2$\times$ compared to SlowFuzz and PerfFuzz. We emphasize that
the slowdown is expected due to functional fitting and clustering
in \toolname.

\noindent\textbf{Worst-case cost of execution}.
We examine the fuzzers outcomes in finding inputs with
the highest cost in terms of executed lines.
Table~\ref{tab:fuzzing-variants} shows that
\toolname finds inputs with higher costs in 5 out of 10 benchmarks
in comparison to SlowFuzz and PerfFuzz.

\noindent\textbf{Discovered paths}.
We consider the number of unique paths discovered by the fuzzers.
A path is unique if it has visited an edge that is not visited by
any other paths. Table~\ref{tab:fuzzing-variants} shows
SlowFuzz has discovered the fewest paths. In 3 out of 10 cases,
PerfFuzz has discovered more paths compared to other fuzzers.

\noindent\textbf{Number of distinct performance functions}.
One requirement of characterizing performance classes is to have
multiple inputs (varied by size) for a path and fit performance functions.
We compare the number of performance functions discovered by the fuzzers.
Table~\ref{tab:fuzzing-variants} shows \toolname finds more performance
functions in 7 out of 10 benchmarks.

\noindent\textbf{Number of functional clusters}.
We consider the number of clusters in performance functions
to show performance classes.
We use the $l_1$ distance between functions for clustering.
We note that the number of clusters should be chosen based on the
quality of clustering for fair comparisons.
As we increase the number of clusters, we measure
the sum of intra-cluster distances as the error.
We pick the optimal number of clusters when
the error is below 1,000 in accordance with standard
elbow method for choosing ideal number of clusters.
Since the error value is the same in all experiments, the
resulting number of clusters is a fair indication of
detecting differential performance.
Table~\ref{tab:fuzzing-variants} shows \toolname finds
more clusters in 4 out of 10 benchmarks.

\begin{figure}[t!]
	\centering
  \begin{minipage}{0.22\textwidth}
  \begin{lstlisting}[caption=InsertionX,label=insertionX]
def insertionX(a,n):
  exchange, i = 0, n-1
  while i > 0:
    if a[i] < a[i-1]:
      swap(arr,i,i-1)
      exchange += 1
    i -= 1
  if exchange == 0:
    return a
  i = 2
  while i < n:
    v, j = a[i], i
    while v < a[j-1]:
      a[j] = a[j-1]
      j -= 1
    a[j], i = v, i+1
  return a
  \end{lstlisting}
	\end{minipage}
  \hfill
	\begin{minipage}{0.25\textwidth}
		\includegraphics[width=\textwidth]{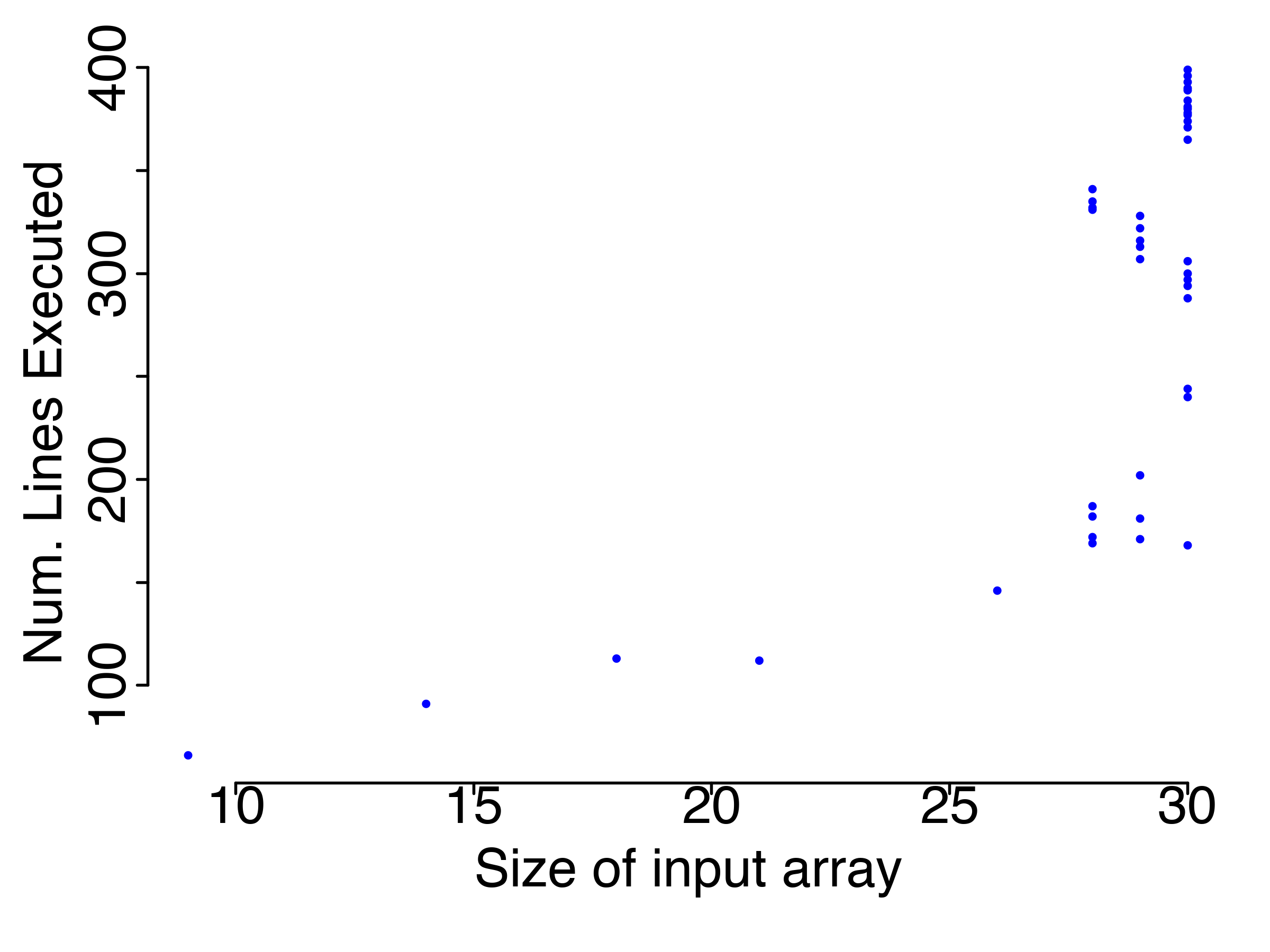}
		\includegraphics[width=\textwidth]{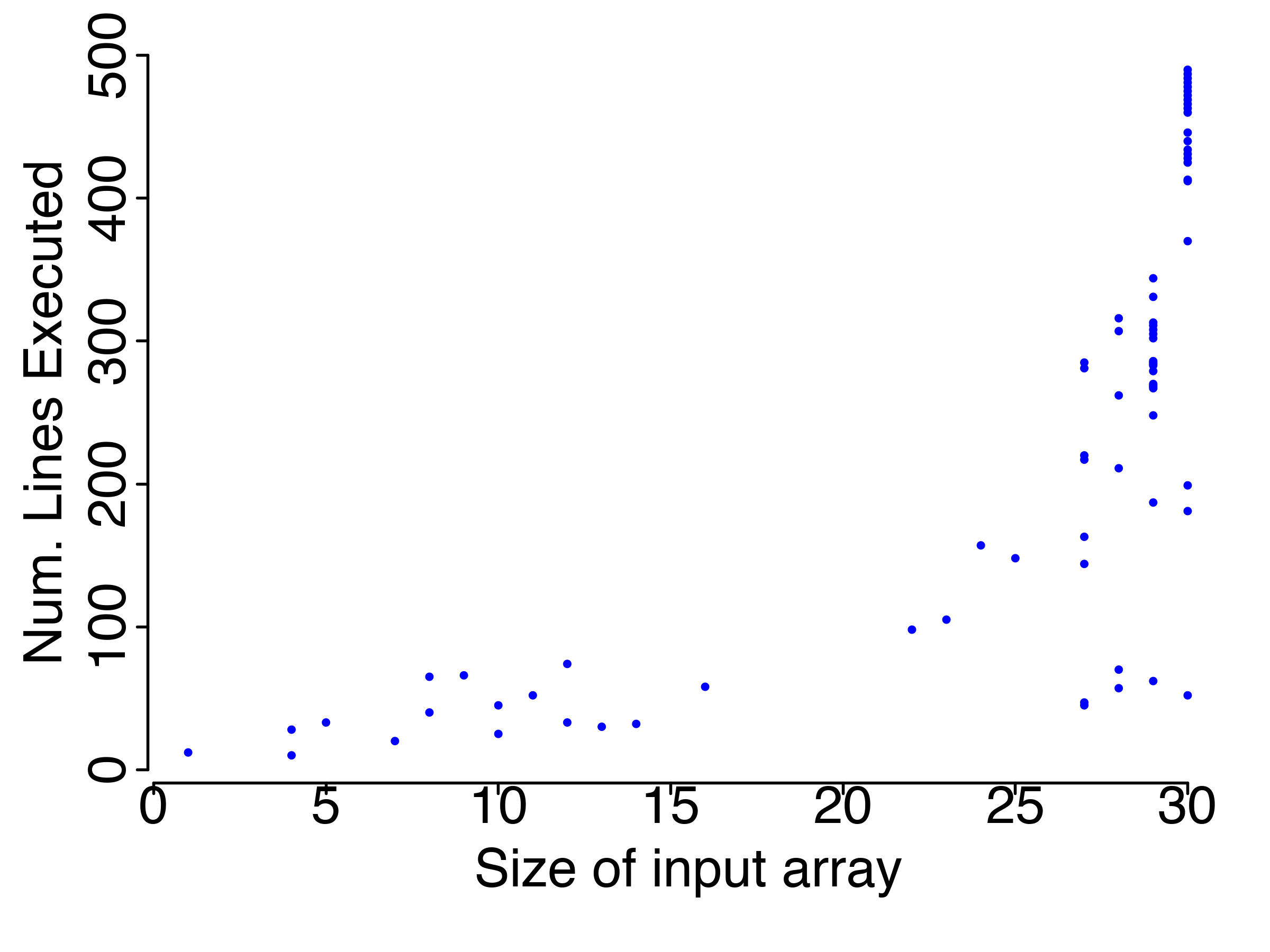}
		\includegraphics[width=\textwidth]{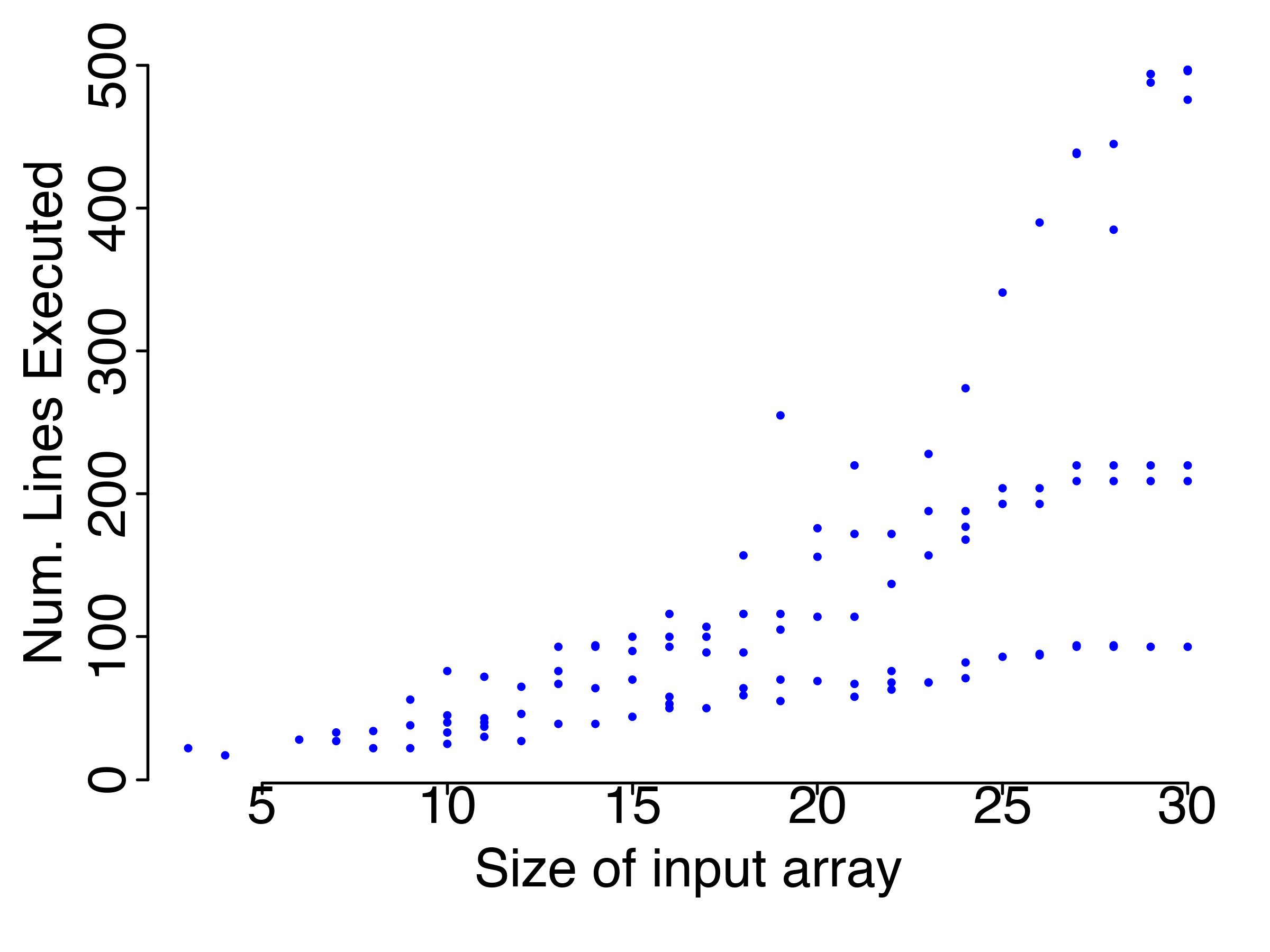}
	\end{minipage}
  \caption{InsertionX. Results
  for SlowFuzz~\cite{petsios2017slowfuzz},
  PerfFuzz~\cite{lemieux2018perffuzz}, and
  \toolname, from top to bottom, respectively.}
	\label{fig:insertionX}
\end{figure}

\noindent\textbf{Summary}. Although \toolname is slower in
generating inputs, it outperforms other fuzzers in
finding worst-case costs, performance
functions, and clusters.
Figure~\ref{fig:insertionX} shows an example of Optimized Insertion sort
(InsertionX). The plots are the performances (in terms of executed lines)
versus the size of inputs for $2*10^6$ samples generated by the fuzzers.

\section{ML Library Analysis}
\label{sec:case-study}
We analyze $8$ larger machine learning (ML) libraries from
scikit-learn~\cite{scikit-learn}.
Although our approach is general enough to apply to any software library
and system (we currently support both Python and Java applications), there are properties
in ML applications that make our approach more practical.
In particular, there is usually a clear distinction between the parameters of the
library and the data in ML libraries. Once we fix the parameters in the given ML
algorithm, the execution times are often a simpler function (e.g. linear and quadratic)
of the number of samples and features in the data.
General programs often do not follow such a structured discipline.
The main research questions
are ``does \toolname (a)
scale well for real-world ML libraries and
(b) provide useful information to debug performance issues?''

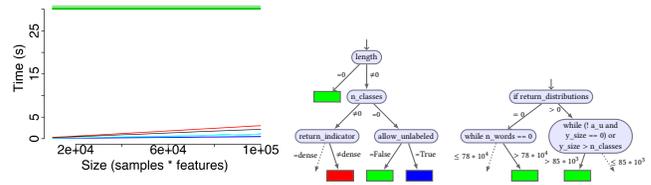
\begin{figure}[t!]
  \centering
  \includegraphics[width=0.2\textwidth]{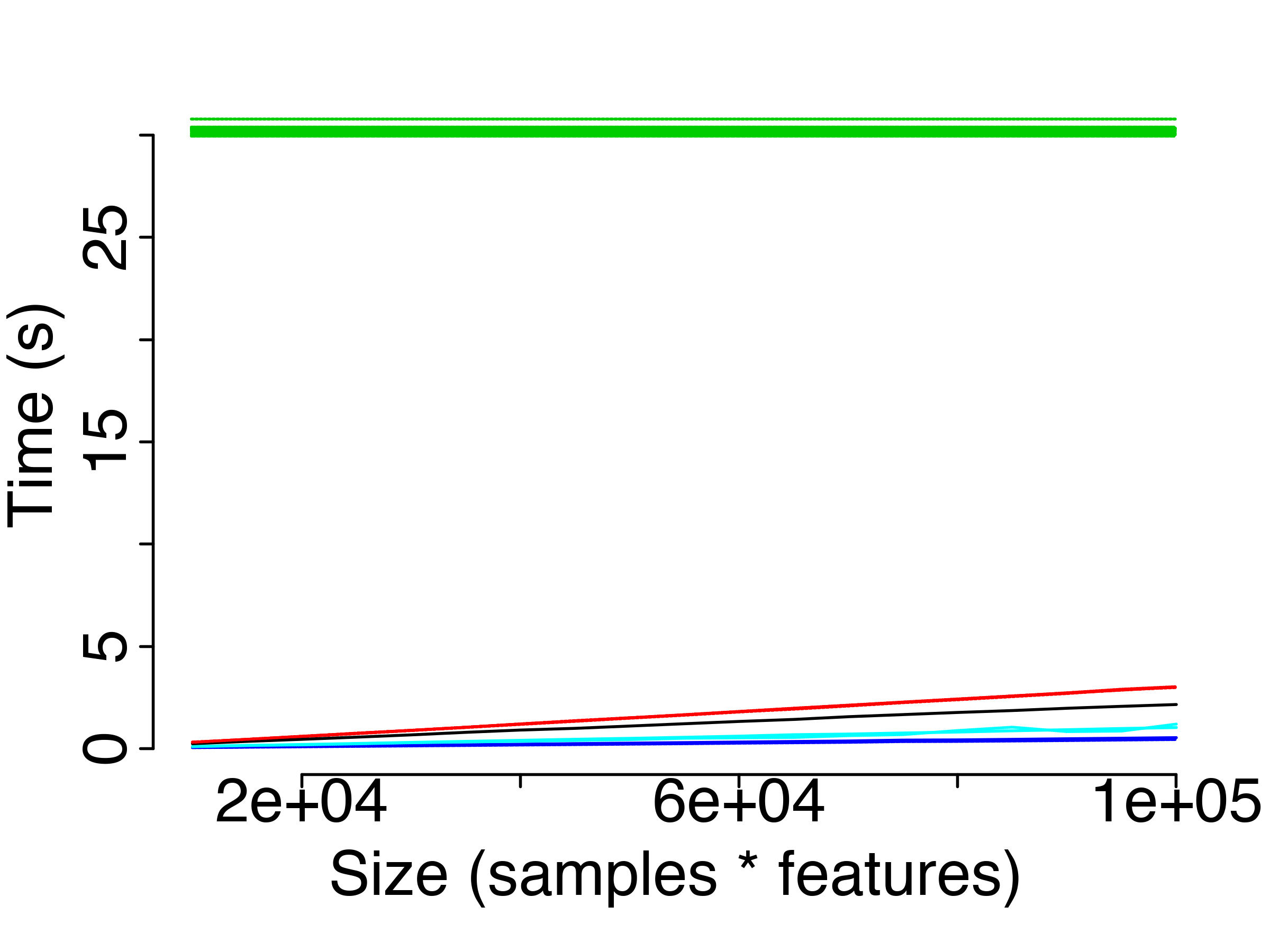}
  \hfill
  \scalebox{0.35}{
    \begin{tikzpicture}[align=center,node distance=1cm,->,thick,
        draw = black!60, fill=black!60]
      \centering
      \pgfsetarrowsend{latex}
      \pgfsetlinewidth{0.3ex}
      \pgfpathmoveto{\pgfpointorigin}
        \node[dtreenode,initial above,initial text={}] at (0,0) (l0)
        {length};
        \node[dtreeleaf,bicolor={green and green and 0.99}] at (-1.5, -1.5)
        (l1) {};
        \node[dtreenode] at (0.0, -1.5) (l2)
        {n\_classes};
        \node[dtreenode] at (1.5, -3.0) (l3)
        {allow\_unlabeled};
        \node[dtreenode] at (-1.5, -3.0) (l4)
        {return\_indicator};
        \node[dtreeleaf,bicolor={green and green and 0.99}] at (0.5, -4.5)
        (l5) {};
        \node[dtreeleaf,bicolor={blue and blue and 0.99}] at (2.0, -4.5)
        (l6) {};
        \node[dtreeleaf,bicolor={red and red and 0.99}] at (-1.0, -4.5)
        (l7) {};
        \node[] at (-2.0, -4.5)
        (l8) {};

        \path[->]  (l0) edge  node [left,pos=0.4] {=0} (l1);
        \path  (l0) edge  node [right, pos=0.4] {$\neq$0} (l2);
        \path  (l2) edge  node [left, pos=0.4] {=0} (l3);
        \path  (l2) edge  node [right, pos=0.4] {$\neq$0} (l4);
        \path  (l3) edge  node [left, pos=0.4] {=False} (l5);
        \path  (l3) edge  node [right, pos=0.4] {=True} (l6);
        \path  (l4) edge  node [right, pos=0.4] {$\neq$dense} (l7);
        \path  (l4) edge[dotted]  node [left, pos=0.4] {=dense} (l8);
    \end{tikzpicture}
    }
    \hfill
    \scalebox{0.35}{
    \begin{tikzpicture}[align=center,node distance=1cm,->,thick,
        draw = black!60, fill=black!60]
      \centering
      \pgfsetarrowsend{latex}
      \pgfsetlinewidth{0.3ex}
      \pgfpathmoveto{\pgfpointorigin}
        \node[dtreenode,initial above,initial text={}] at (0,0) (l0)
        {if return\_distributions};
        \node[dtreenode] at (-2.0, -1.5) (l1)
        {while n\_words == 0};
        \node[dtreenode] at (1.5, -1.5) (l2)
        {while ($!$ a\_u and \\y\_size == 0) or\\ y\_size $>$ n\_classes};
        \node[dtreeleaf,bicolor={green and green and 0.99}] at (-1.0, -3.0)
        (l3) {};
        \node[dtreeleaf,bicolor={green and green and 0.99}] at (1.0, -3.0)
        (l4) {};
        \node[] at (-2.5, -3.0)
        (l5) {};
        \node[] at (2.5, -3.0)
        (l6) {};
        \path[->]  (l0) edge  node [left,pos=0.4] {$= 0$} (l1);
        \path  (l0) edge  node [left,pos=0.4] {$> 0$} (l2);
        \path  (l1) edge[dotted]  node [left, pos=0.4] {$\leq 78*10^4$} (l5);
        \path  (l1) edge  node [right, pos=0.4] {$> 78*10^4$} (l3);
        \path  (l2) edge  node [left, pos=0.4] {$> 85*10^3$} (l4);
        \path  (l2) edge[dotted]  node [right, pos=0.4] {$\leq 85*10^3$} (l6);
    \end{tikzpicture}
  }
  \caption{
    (a) Inputs of make classification are clustered into 5 groups.
    (b) Decision tree model based on input parameter features.
    (c) Decision tree model based on internal features.
  }
  \label{fig:make-multiclass-data-clustered}
  \label{fig:make-multiclass-data-input}
  \label{fig:make-multiclass-data-internal}
\end{figure}

\noindent\textbf{A) Logistic Regression Classifier.}
In summary, \toolname
detects $4$ clusters after fuzzing for about $2$ hours. The debugging
revealed a performance bug in the implementation of logistic regression
that has since been fixed (see details in Overview section~\ref{sec:overview}).

\noindent\textbf{B) Make Classification Data Set Util.}
We analyze the performance of \texttt{make\_multilabel\_classification}
method inside \texttt{samp le\_generator} module~\cite{make-multilabel-classification}.

\noindent\textit{Fuzzing and clustering}. \toolname provides $243$
sets of inputs related to different paths in the module after running
for $4$ hours. The fuzzing covers $293$ LoC from almost $350$ LoC in the
implementations of this method and its dependencies.
Figure~\ref{fig:make-multiclass-data-clustered} (a) shows that the
$243$ performance functions are clustered into $5$ groups.
The clustering
shows a huge differential performance between the green cluster and other clusters.

\noindent\textit{Analyzing input space}.
Figure~\ref{fig:make-multiclass-data-clustered} (b) shows that
if the length parameter sets to 0, then the input is in the
green cluster. Another way to see
the expensive green cluster is to set the value of n\_class parameter
to 0 and allow\_unlabeled parameter to False.

\noindent\textit{Bug localization}.
The decision tree model in
Figure~\ref{fig:make-multiclass-data-clustered} (c) shows
the green cluster is associated with the calls to two different
loops. The following code snippet shows these parts
in \texttt{make multilabel classification} module:

\vspace{0.5em}
\begin{scriptsize}
\begin{lstlisting}
def sample_example():
  ...
  while (not allow_unlabeled and
    y_size == 0) or y_size > n_classes:
      y_size = generator.poisson(n_labels)
  ...
  while n_words == 0:
      n_words = generator.poisson(length)
  ...
\end{lstlisting}
\end{scriptsize}

The code snippet shows the possibility of an infinite number of executions
for the two loop bodies
(the fuzzer terminates processes if the execution of inputs takes more than
$15$ minutes).
We mark these two parts in \texttt{make\_multilabel\_ classification}
method performance bugs \circled{2},\circled{3}. We have reported these
issues to scikit-learn developers~\cite{make-classification-dataset-bug}.
They confirmed the bugs and have since
fixed them~\cite{make-classification-dataset-fix}.
The fix does not allow parameters n\_classes and length to
accept zero values. There are also two division by zero crashes in
\texttt{make\_classification} method discovered during fuzzing if
the n\_classes parameter or n\_clusters\_per \_class parameter set to zero:

\vspace{0.5em}
\begin{scriptsize}
\begin{lstlisting}
  ...
  weights = [1.0 / n_classes] * n_classes
  ...
  weights[k % n_c] / n_clusters_per_class
\end{lstlisting}
\end{scriptsize}

\noindent\textit{Scalability.} In 240 minutes, \toolname generates
243 performance functions. For debugging,
\toolname generates 293 internal features and uses
the features to learn the decision tree in 1(s).

\noindent\textit{Usefulness.} \toolname discovers and pinpoints
2 performance bugs and 2 division by zero crashes in the
implementations of \texttt{make\_multi label\_classification} method.

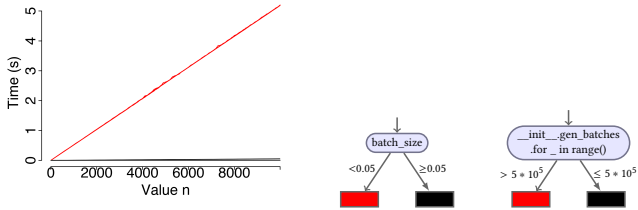
\begin{figure}[t!]
  \centering
  \includegraphics[width=0.22\textwidth]{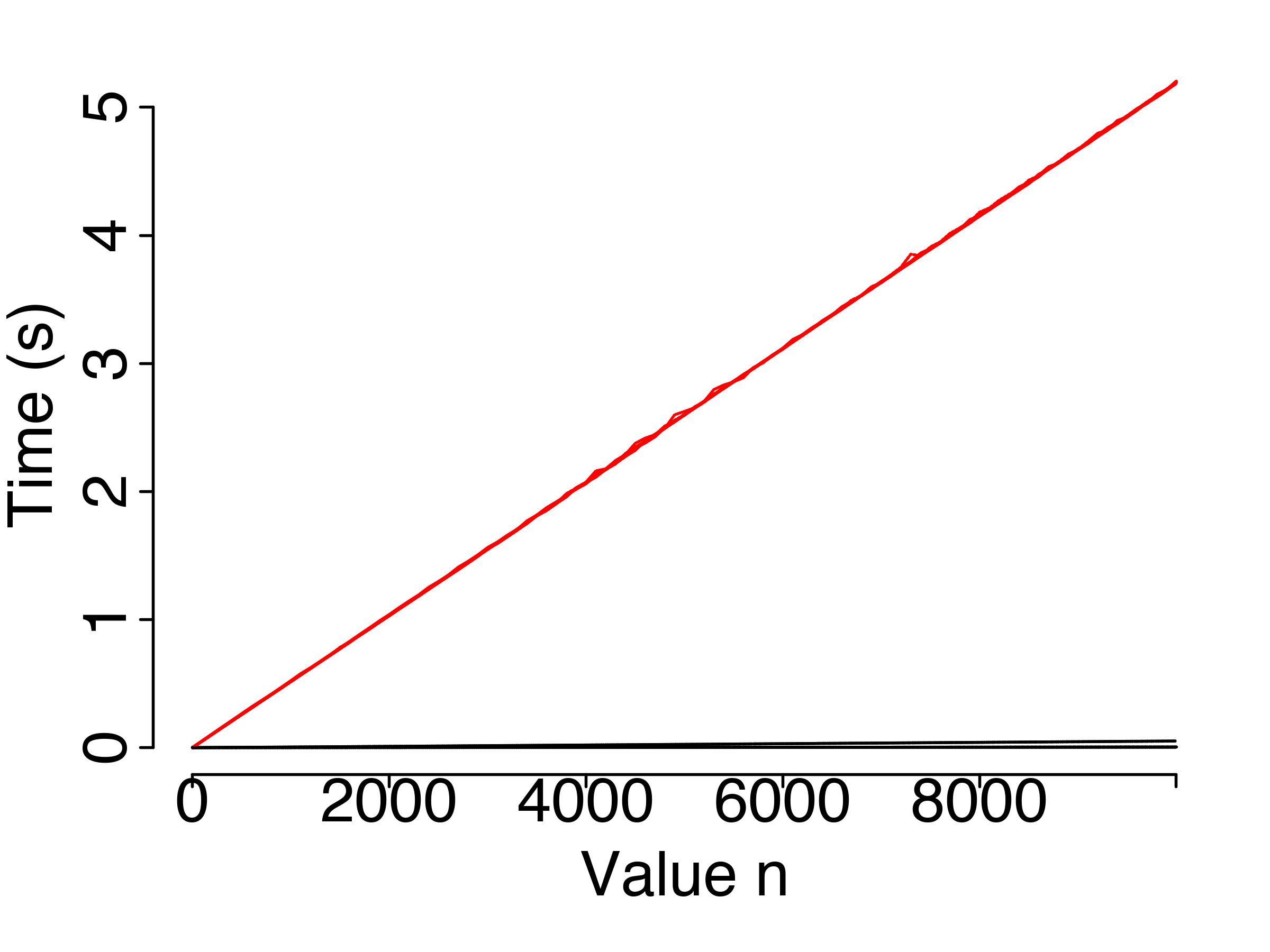}
  \hfill
  \scalebox{0.5}{
    \begin{tikzpicture}[align=center,node distance=1cm,->,thick,
        draw = black!60, fill=black!60]
      \centering
      \pgfsetarrowsend{latex}
      \pgfsetlinewidth{0.3ex}
      \pgfpathmoveto{\pgfpointorigin}
        \node[dtreenode,initial above,initial text={}] at (0,0) (l0)
        {batch\_size};
        \node[dtreeleaf,bicolor={red and red and 0.99}] at (-1.0, -1.5)
        (l1) {};
        \node[dtreeleaf,bicolor={black and black and 0.99}] at (1.0, -1.5)
        (l2) {};

        \path[->]  (l0) edge  node [left,pos=0.4] {$<$0.05} (l1);
        \path  (l0) edge  node [right, pos=0.4] {$\geq$0.05} (l2);
    \end{tikzpicture}
    }
    \hfill
    \scalebox{0.5}{
    \begin{tikzpicture}[align=center,node distance=1cm,->,thick,
        draw = black!60, fill=black!60]
      \centering
      \pgfsetarrowsend{latex}
      \pgfsetlinewidth{0.3ex}
      \pgfpathmoveto{\pgfpointorigin}
        \node[dtreenode,initial above,initial text={}] at (0,0) (l0)
        {\_\_init\_\_.gen\_batches\\.for \_ in range()};
        \node[dtreeleaf,bicolor={red and red and 0.99}] at (-1.0, -1.5)
        (l1) {};
        \node[dtreeleaf,bicolor={black and black and 0.99}] at (1.0, -1.5)
        (l2) {};

        \path[->]  (l0) edge  node [left,pos=0.4] {$ > 5*10^5$} (l1);
        \path  (l0) edge  node [right, pos=0.4] {$\leq 5*10^5$} (l2);
    \end{tikzpicture}
  }
  \caption{
    (a) Inputs generated by \toolname for \texttt{util} module
    clustered.
    (b) Decision tree using the input parameters of \texttt{util}.
    (c) Decision tree using the internal features of \texttt{util}.
  }
  \label{fig:gen-batch-clustered}
  \label{fig:gen-batch-classified-input}
  \label{fig:gen-batch-classified-internal}
\end{figure}

\noindent\textbf{C) Batch Generator.}
We analyze the implementation of
\texttt{util}\footnote{\url{https://github.com/scikit-learn/scikit-learn/blob/master/sklearn/utils/__init__.py}}
in~\cite{scikit-learn}.
This module provides various utilities such as generating slices of certain
sizes for the given data.

\noindent\textit{Fuzzing and clustering}.
We fuzz batch generator methods of this module
for $30$ minutes and obtain $20$ sets of inputs.
Figure~\ref{fig:gen-batch-clustered} (a) shows that
there are two clusters of performance.

\noindent\textit{Analyzing input space}.
Figure~\ref{fig:gen-batch-classified-input} (b)
shows that the differences between
red and black patterns are related to batch\_size parameter
and the library accepts small positive float values for this parameter
such as $0.001$.

\noindent\textit{Bug localization}.
Figure~\ref{fig:gen-batch-classified-internal} (c)
pinpoints the following loop body in \texttt{gen\_batches()}:

\vspace{0.5em}
\begin{scriptsize}
\begin{lstlisting}
  for _ in range(int(n // batch_size)):
    end = start + batch_size
    ...
\end{lstlisting}
\end{scriptsize}

The decision tree shows that the loop above can be taken
millions of times if the batch\_size sets to small positive values close to $0$.
We mark this as performance bug~\circled{4}. This bug
has since been confirmed and fixed by the
developers~\cite{batch-generator-bug,batch-generator-fix}. The fix
checks the batch\_size parameter to be both integer and greater than or equal to 1.

\noindent\textit{Scalability.} During $30$ mins of fuzzing,
\toolname generates $20$ performance functions.
\toolname generates $15$ internals features
and uses the features to infer the decision tree in 0.1(s).

\noindent\textit{Usefulness.} \toolname discovers and pinpoints a
performance bug in the \texttt{util} module.

\begin{figure}[t!]
\centering
\includegraphics[width=0.22\textwidth]{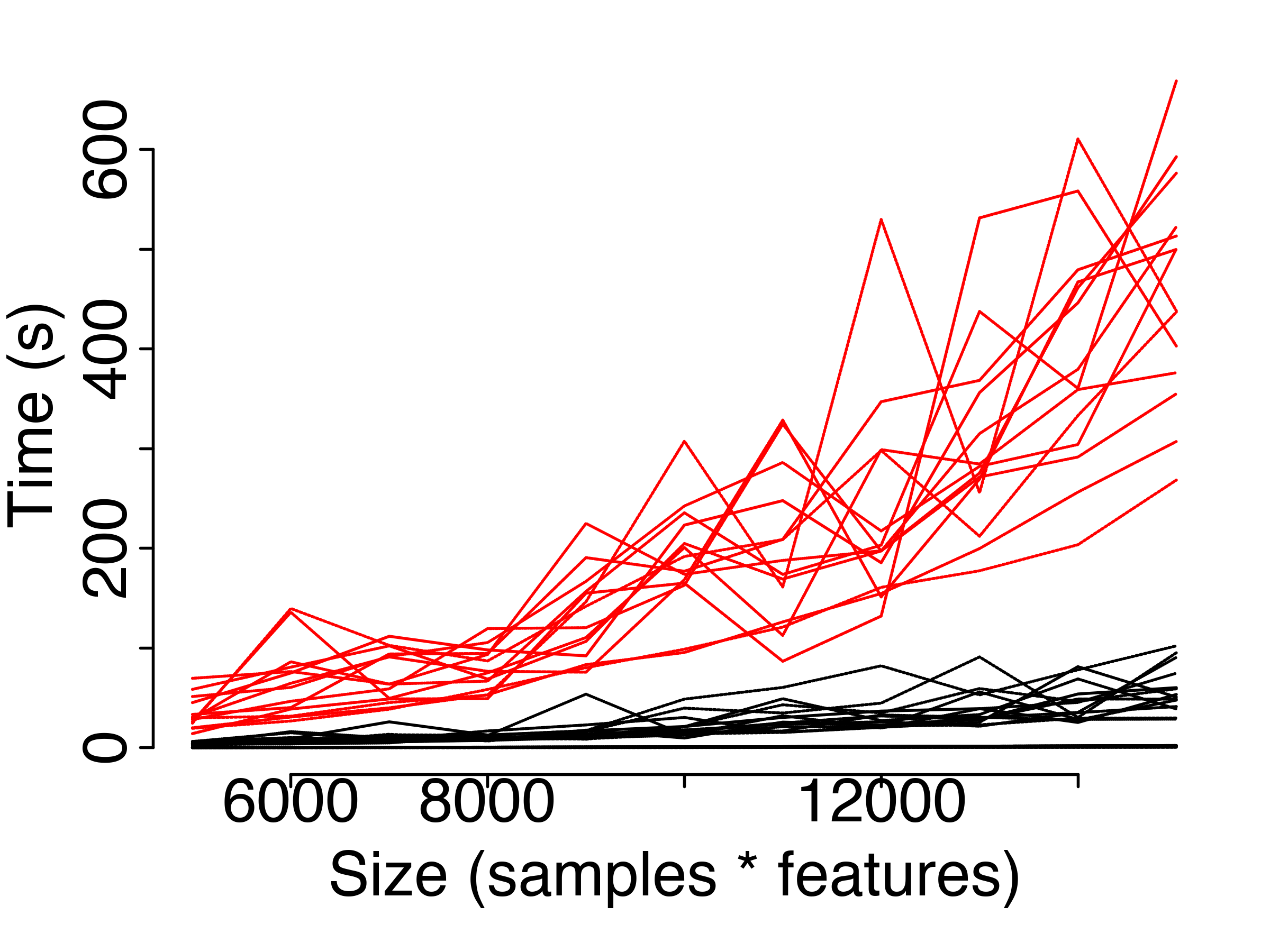}
\hfill
\scalebox{0.45}{
  \begin{tikzpicture}[align=center,node distance=1cm,->,thick,
      draw = black!60, fill=black!60]
    \centering
    \pgfsetarrowsend{latex}
    \pgfsetlinewidth{0.3ex}
    \pgfpathmoveto{\pgfpointorigin}
      \node[dtreenode,initial above,initial text={}] at (0,0) (l0)
      {optimizer};
      \node[dtreeleaf,bicolor={black and black and 0.99}] at (-1.0, -1.5)
      (l1) {};
      \node[dtreenode] at (1.0, -1.5) (l2)
      {n\_restarts\_\\optimizer};
      \node[dtreeleaf,bicolor={red and red and 0.99}] at (0.0, -3.0)
      (l3) {};
      \node[dtreeleaf,bicolor={black and black and 0.99}] at (2.0, -3.0)
      (l4) {};

      \path[->]  (l0) edge  node [left,pos=0.4] {=None} (l1);
      \path  (l0) edge  node [right, pos=0.4] {=fmin\_l\_\\bfgs\_b} (l2);
      \path  (l2) edge  node [left, pos=0.4] {$>$6} (l3);
      \path  (l2) edge  node [right, pos=0.4] {$\leq$6} (l4);
  \end{tikzpicture}
  }
  \hfill
  \scalebox{0.45}{
  \begin{tikzpicture}[align=center,node distance=1cm,->,thick,
      draw = black!60, fill=black!60]
    \centering
    \pgfsetarrowsend{latex}
    \pgfsetlinewidth{0.3ex}
    \pgfpathmoveto{\pgfpointorigin}
      \node[dtreenode,initial above,initial text={}] at (0,0) (l0)
      {scipy.optimize.lbfgsb.\\while 1};
      \node[dtreeleaf,bicolor={red and red and 0.99}] at (-1.0, -1.5)
      (l1) {};
      \node[dtreeleaf,bicolor={black and black and 0.99}] at (1.0, -1.5)
      (l2) {};
      \path[->]  (l0) edge  node [left,pos=0.4] {$> 0$} (l1);
      \path  (l0) edge  node [right, pos=0.4] {$\leq$ 0} (l2);
  \end{tikzpicture}
}

\caption{
  (a) Inputs generated by \toolname for \texttt{Gaussian} classifier
  clustered.
  (b) Decision tree using the input parameters of \texttt{Gaussian}.
  (c) Decision tree using the internal features of \texttt{Gaussian}.
}
\label{fig:gau-process-clustered}
\label{fig:gau-process-classified-input}
\label{fig:gau-process-classified-internal}
\end{figure}
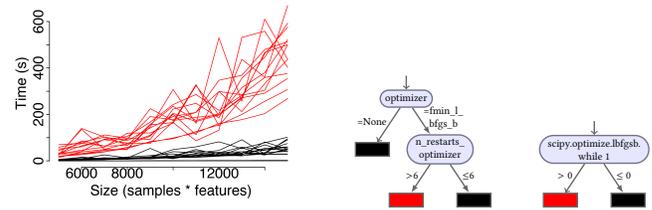

\noindent\textbf{D) Gaussian Process Classification.}
We analyze the implementations of Gaussian Process (GP) as a classifier
model~\cite{gaussian-process-implementations}
in the scikit-learn library~\cite{scikit-learn}. This classifier is specifically used
for probabilistic classification (see~\cite{gaussian-process-explanation} for more details
about the functionality of this classifier). We fix non-deterministic
(stochastic) behaviors in the library to a deterministic random value.

\textit{Fuzzing and clustering}.
We run \toolname for $240$ minutes and obtain $173$ sets of inputs.
Figure~\ref{fig:gau-process-clustered} (a)
shows a huge performance difference between the black and red
clusters (more than $10$ times).

\noindent\textit{Analyzing input space}.
Figure~\ref{fig:gau-process-classified-input} (b) shows that if the
optimizer parameter sets to `fmin\_l\_bfgs\_b' and n\_restarts\_optimizer parameter
sets to values more than $6$, then the input is in the (slow) red cluster.
Otherwise, the input follows the (fast) black cluster.

\noindent\textit{Bug localization}.
Figure~\ref{fig:gau-process-classified-internal} (c) shows
that the number of calls to the loop body inside
\texttt{scipy.optimize.lbfgsb} module causes the performance
differences:

\vspace{0.5em}
\begin{scriptsize}
\begin{lstlisting}
while 1:
  ...
  _lbfgsb.setulb(...
  ...
\end{lstlisting}
\end{scriptsize}

It seems that the Gaussian classifier runs for the number of
n\_restarts\_optimizer parameter and it causes huge performance
differences by calling to \texttt{lbfgsb} optimizer.
Since the behavior is related to external library, we left
further analysis for future work to determine if the behavior is intrinsic
to the problem, or it is a performance bug.

\noindent\textit{Scalability.} During $240$ mins, \toolname
obtains $173$ performance functions. \toolname generates
$1,333$ features about program internals and uses the
features to learn the decision tree model in 5(s).

\noindent\textit{Usefulness.} \toolname discovers and
pin-points a code region in an external library (scipy optimizer).

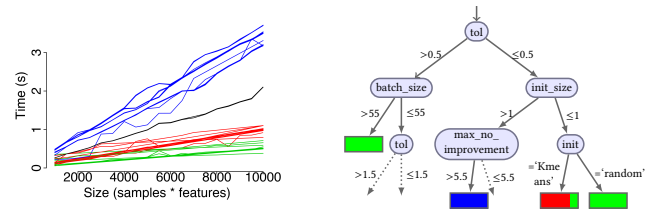
\begin{figure}[t!]
  \centering
  \includegraphics[width=0.2\textwidth]{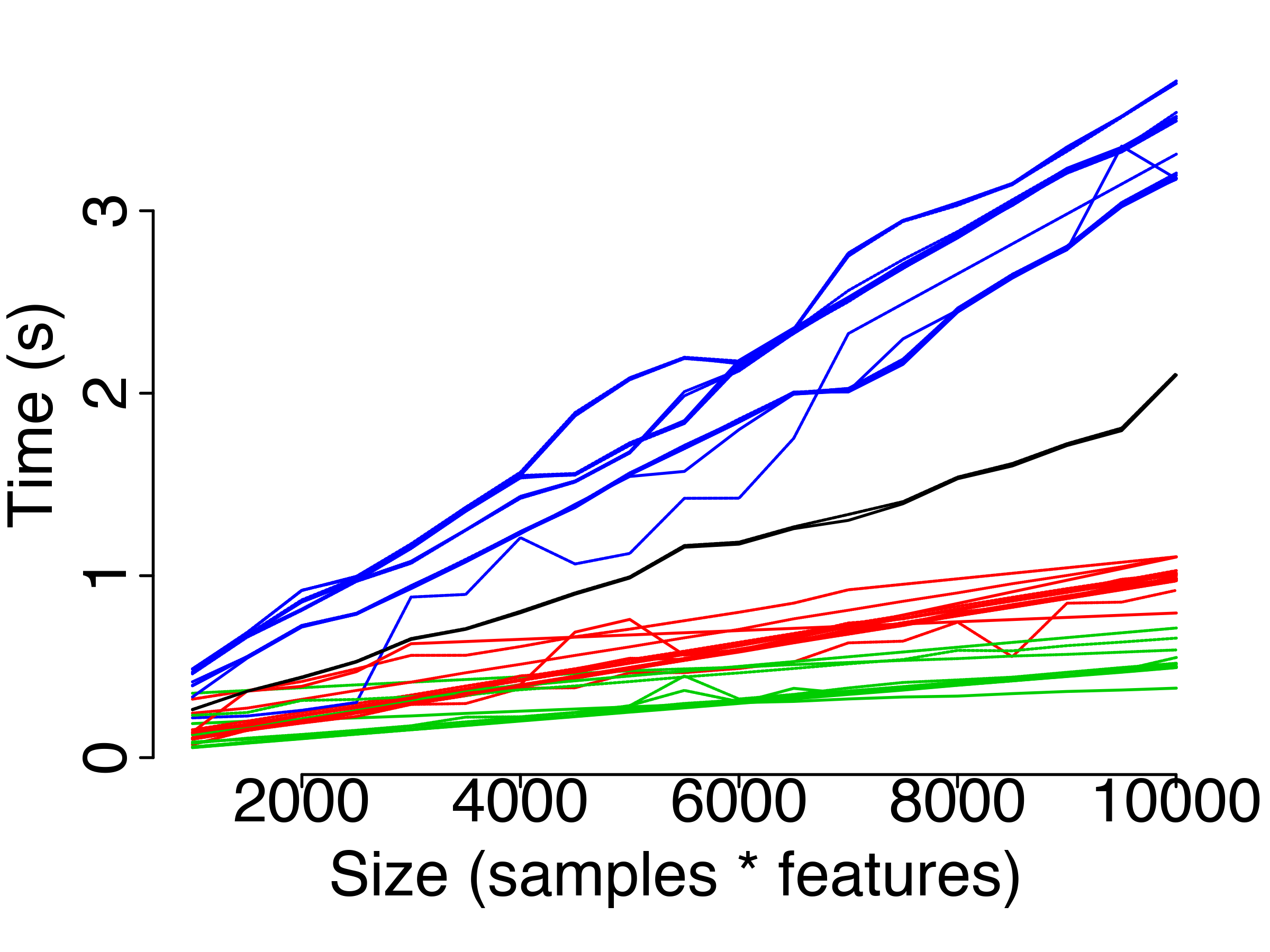}
  \hfill
  \scalebox{0.5}{
    \begin{tikzpicture}[align=center,node distance=1cm,->,thick,
        draw = black!60, fill=black!60]
      \centering
      \pgfsetarrowsend{latex}
      \pgfsetlinewidth{0.3ex}
      \pgfpathmoveto{\pgfpointorigin}
        \node[dtreenode,initial above,initial text={}] at (0,0) (l0)
        {tol};
        \node[dtreenode] at (-2,-1.5) (l1)
        {batch\_size};
        \node[dtreenode] at (2, -1.5) (l2)
        {init\_size};
        \node[dtreeleaf,bicolor={green and green and 0.99}] at (-3.0, -3) (l3)
        {};
        \node[dtreenode] at (-2.0, -3) (l4)
        {tol};
        \node[dtreenode] at (0.0, -3) (l5)
        {max\_no\_\\improvement};
        \node[dtreenode] at (2.5, -3) (l6)
        {init};
        \node at (-3.0, -4.5) (l7)
        {};
        \node at (-2.0, -4.5) (l8)
        {};
        \node[dtreeleaf,bicolor={blue and blue and 0.99}] at (-0.2, -4.5)
        (l9) {};
        \node at (0.5, -4.5)
        (l10) {};
        \node[dtreeleaf,bicolor={red and green and 0.8}] at (2.2, -4.5)
        (l11) {};
        \node[dtreeleaf,bicolor={green and green and 0.99}] at (3.5, -4.5)
        (l12) {};

        \path[->]  (l0) edge  node [left,pos=0.4] {$>$0.5} (l1);
        \path  (l0) edge  node [right, pos=0.4] {$\leq$0.5} (l2);
        \path  (l1) edge  node [left, pos=0.4] {$>$55} (l3);
        \path  (l1) edge node [right,pos=0.4] {$\leq$55} (l4);
        \path  (l2) edge  node [left] {$>$1} (l5);
        \path  (l2) edge  node [right] {$\leq$1} (l6);
        \path  (l4) edge[dotted]  node [left] {$>$1.5} (l7);
        \path  (l4) edge[dotted]  node [right] {$\leq$1.5} (l8);
        \path  (l5) edge  node [left] {$>$5.5} (l9);
        \path  (l5) edge[dotted]  node [right] {$\leq$5.5} (l10);
        \path  (l6) edge  node [left] {=`Kme\\ans'} (l11);
        \path  (l6) edge  node [right] {=`random'} (l12);
    \end{tikzpicture}
    }
  \caption{
    (a) Inputs generated by~\toolname for mini-batch \texttt{KMeans} are clustered
    into 4 groups.
    (b) Decision tree using the input parameters of \texttt{KMeans}.
    %
    %
  }
  \label{fig:minibatch-kmeans-clustered}
  \label{fig:minibatch-kmeans-classified-input}
\end{figure}

\noindent\textbf{E) Mini-batch KMeans.}
We analyze the implementations of KMeans mini-batch
clustering~\cite{minibatch-kmeans}. This is a variant of the KMeans algorithm
which uses mini-batches to reduce the computation time, while still attempting
to optimize the same objective function. We assume that the
initial cluster centroids are deterministically chosen by setting
seed values to a constant value.

During a run of $240$ mins, we obtain $325$ input
sets from \toolname.
Figure~\ref{fig:minibatch-kmeans-clustered} (a) shows that
$325$ functions are clustered in $4$ groups.
Figure~\ref{fig:minibatch-kmeans-classified-input} (b) shows
the expensive blue cluster happens if the `tol' is less than or equal
to $0.5$, the `init\_size' is larger than or equal to $1$, and the
`max\_no\_improvement' is larger than $5.5$.
Looking into the internal features,
we observe that the number of calls to calculate the euclidean
distance between points and the squared differences
between the current and the previous errors
are important discriminant features. However,
the discriminant features seem to explain behaviors
that are intrinsic to the problem.

\begin{figure}[t!]
  \centering
  \includegraphics[width=0.25\textwidth]{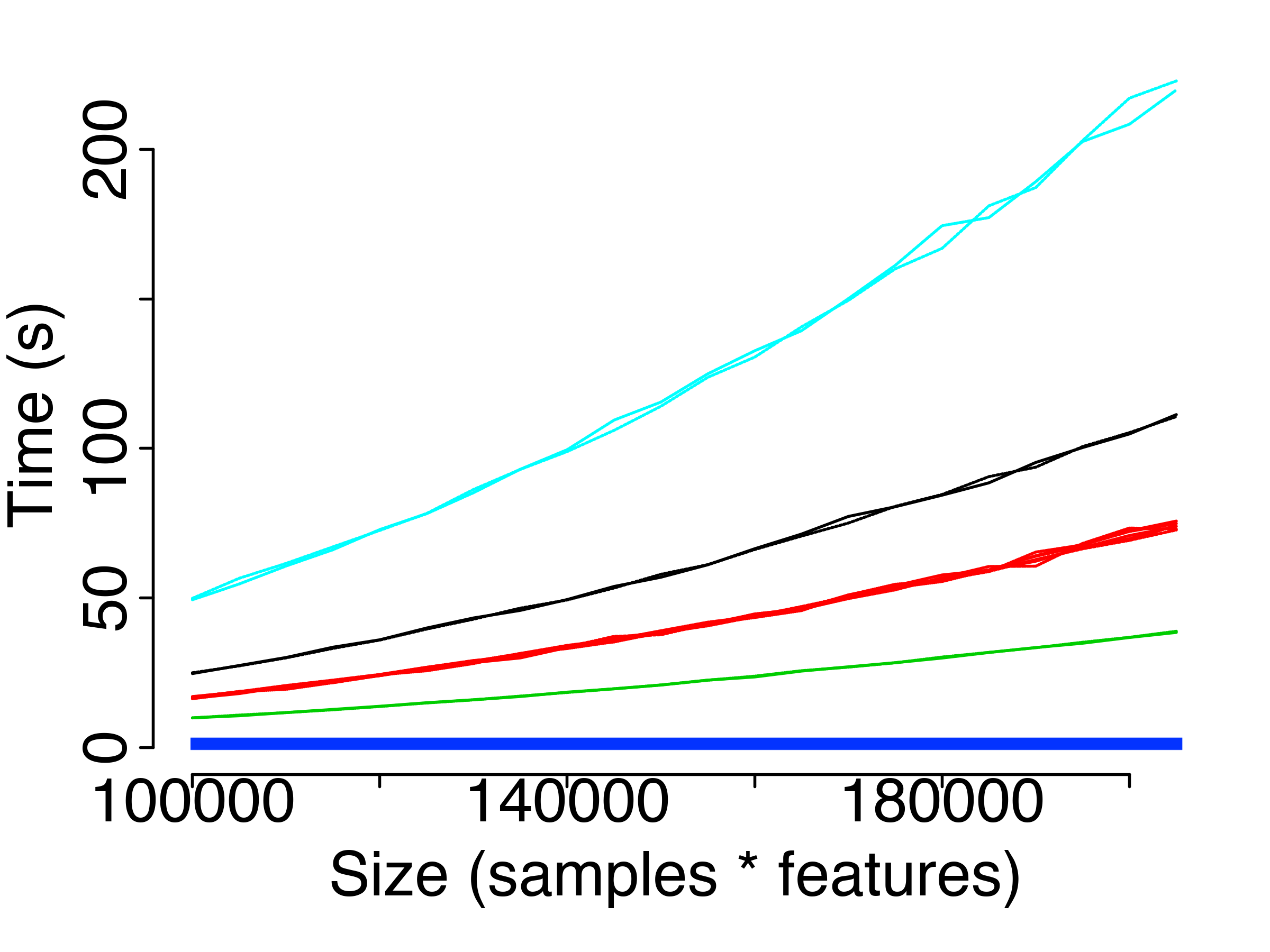}
  \hfill
  \scalebox{0.5}{
    \begin{tikzpicture}[align=center,node distance=1cm,->,thick,
        draw = black!60, fill=black!60]
      \centering
      \pgfsetarrowsend{latex}
      \pgfsetlinewidth{0.3ex}
      \pgfpathmoveto{\pgfpointorigin}
        \node[dtreenode,initial above,initial text={}] at (0,0) (l0)
        {bootstrap};
        \node[dtreenode] at (-2,-1.5) (l1)
        {criterion};
        \node[dtreenode] at (2, -1.5) (l2)
        {n\_estimators};
        \node[dtreenode] at (-3.0, -3) (l3)
        {max\_features};
        \node[dtreeleaf,bicolor={blue and blue and 0.99}] at (-1.0, -3)
        (l4) {};
        \node[dtreenode] at (1.0, -3) (l5)
        {criterion};
        \node[dtreeleaf,bicolor={blue and blue and 0.99}] at (2.5, -3)
        (l6) {};
        \node[dtreenode] at (-3.0, -4.5) (l7)
        {max\_depth};
        \node[dtreeleaf,bicolor={red and red and 0.99}] at (-1.0, -4.5)
        (l8) {};
        \node[dtreeleaf,bicolor={green and green and 0.99}] at (1.0, -4.5)
        (l9) {};
        \node[dtreeleaf,bicolor={blue and blue and 0.99}] at (2.5, -4.5)
        (l10) {};
        \node[dtreeleaf,bicolor={black and black and 0.99}] at (-3.0, -6.0)
        (l11) {};
        \node[dtreeleaf,bicolor={cyan and cyan and 0.99}] at (-1.0, -6.0)
        (l12) {};

        \path[->]  (l0) edge  node [left,pos=0.4] {=False} (l1);
        \path  (l0) edge  node [right, pos=0.4] {=True} (l2);
        \path  (l1) edge  node [left, pos=0.4] {=`mae'} (l3);
        \path  (l1) edge node [right,pos=0.4] {=`mse'} (l4);
        \path  (l2) edge  node [left] {$>$112} (l5);
        \path  (l2) edge  node [right] {$\leq$112} (l6);
        \path  (l3) edge  node [left] {= `auto'} (l7);
        \path  (l3) edge  node [right] {=`sqrt'} (l8);
        \path  (l5) edge  node [left] {=`mae'} (l9);
        \path  (l5) edge  node [right] {=`mse'} (l10);
        \path  (l7) edge  node [left] {=`None'} (l11);
        \path  (l7) edge  node [right] {=`rndInt'} (l12);

    \end{tikzpicture}
  }

  \caption{
    (a) Inputs generated by~\toolname for forest \texttt{regressor}
    are clustered into 5 groups.
    (b) Decision tree using the input features of forest \texttt{regressor}.
  }
  \label{fig:forest-regression-clustered}
  \label{fig:forest-regression-classified}
\end{figure}
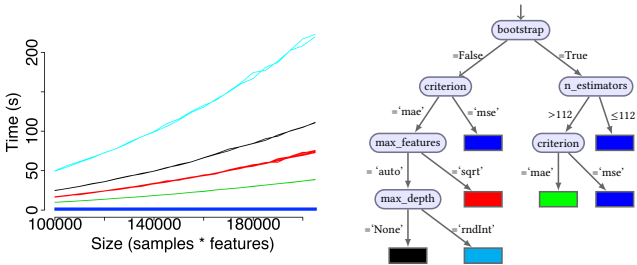

\noindent\textbf{F) Random Forest Regressor.}
A random forest~\cite{random-forest-regressor} is a meta estimator that
fits a number of classifying
decision trees on various sub-samples of the dataset and
uses averaging to improve the predictive accuracy and control over-fitting.
We analyze its implementations in scikit-learn library~\cite{scikit-learn}.

We obtain $160$ sets of inputs from \toolname after running for $240$ mins.
Figure~\ref{fig:forest-regression-clustered} (a) shows (subset of)
inputs are clustered into 5 groups.
The clustering shows that the computational complexity in the random
forest regressor can be quadratic.
The decision tree model in Figure~\ref{fig:forest-regression-classified} (b)
shows that the inputs with criterion parameter sets
to `mae' have higher costs.
A similar issue has already reported as a potential
performance bug in the regressor~\cite{rnd-forest-issue}.
Learning forest regressors is expensive with `mae' criteria
since it sorts inputs in each step of
learning to calculate the median.
Since the root causes seem to be related to external library, we left
further analysis for future work to determine if the behavior is intrinsic
to the problem, or it is a performance bug.

\noindent\textbf{G) Discriminant Analysis.}
The discriminant analysis~\cite{discriminant-analysis-lda-qda}
is a classic classifier with linear and quadratic
decision boundaries. We analyze linear and quadratic discriminant analysis
implemented in scikit-learn~\cite{scikit-learn}.
During 52 mins of fuzzing, we obtain 78 sets of inputs.
Upon clustering, we observe
that the point-wise distances between clusters are in order of
fractions of a second. Therefore, we gain more confidence that
the discriminant analysis is free of performance issues.

\noindent\textbf{H) Decision Tree Classifier.}
The decision tree classifier~\cite{decision-tree-scikit-learn} is a white-box
classification model that predicts the target variable by inferring decision rules.
We analyze the implementations of this model in scikit-learn~\cite{scikit-learn}.
After fuzzing for 240 mins, \toolname generates 492 sets of inputs.
Upon clustering, we realize the point-wise distances between centroids are
less than 0.1 second. Thus, we become more confident
that the decision tree classifier is free of performance bugs.

\section{Related Work}
\label{sec:related}
\noindent\textbf{Performance Fuzzing}.
Evolutionary algorithms have been widely used for finding inputs
that trigger the worst-case
complexity~\cite{petsios2017slowfuzz,lemieux2018perffuzz}.
SlowFuzz~\cite{petsios2017slowfuzz} extends the libFuzzer~\cite{libFuzzer}
to discover DoS bugs, that is, inputs with expensive
computations such as exponential. In contrast, \toolname
is looking for different classes of computational
complexities rather than only the worst-case one. PerfFuzz~\cite{lemieux2018perffuzz}
is the closest fuzzing technique to us. The goal is to maximize
the cost of different entities in the program (edges in CFG)
that help characterize the worst-case behaviors in large-scale systems.
PerfFuzz has a single global population whereas \toolname has multiple populations,
one population per path. This model of population in \toolname enables the debugger
to model performance functions precisely and obtain diverse classes of performances.
In addition, \toolname utilizes clustering during fuzzing
to discover a few diverse performance classes rather than
all entity classes, many of those may have similar performance.

\noindent\textbf{Differential Fuzzing}.
DifFuzz~\cite{DBLP:conf/icse/nilizadeh} has developed on top of
AFL~\cite{AFL} and Kelinci~\cite{kersten2017poster} to discover
information leaks due to timing side channels in Java programs.
DifFuzz adapts the traditional notion of confidentiality, noninterference.
A program is \textit{unsafe} iff for a pair of secret values
$s_1$ and $s_2$, there exists a public value $p$ such that the behavior of the
program on $(s_1,p)$ is observably different than on $(s_2,p)$.
The goal of DifFuzz is to maximize the following objective:
${\delta = |c(p,s_1) - c(p,s_2)|}$,
that is, to find two distinct secret values $s_1,s_2$ and a public value $p$
that give the maximum cost ($c$) difference in two runs of a program.
If the difference of any pair of secret values is more than $\epsilon$,
the program is considered to be vulnerable to timing side-channel attacks.
In contrast to DifFuzz, we are looking for $k$ sets of inputs
to discover differential performances in python-based machine learning
libraries. Fuchsia~\cite{FuncSideChan18} is another technique that performs
differential analysis to debug timing side channels. For detection,
Fuchsia extends AFL to characterize response times as functions over
public inputs. Fuchsia performs fuzzing and clustering in two separate
steps, wheres DPFuzz combines these two steps to explore the space of
input efficiently. In addition, the time model in Fuchsia can be in
arbitrary shapes, while the performance model in DPFuzz often follows simpler
shapes such as polynomials.

\noindent\textbf{Debugging Performances}.
Machine learning and statistical models have been used for
fault localization~\cite{wong2016survey} and debugging of performance
issues~\cite{song2014statistical,aaai18,tizpaz2017discriminating}.
These works generally assume that the interesting inputs are given, while
we adapt evolutionary-based fuzzing techniques to automatically generate interesting
inputs. DPDEBUGGER~\cite{aaai18} considers a model of programs
where the inputs are not gathered as functional data.
Therefore, it needs to discover performance
functions. DPDEBUGGER extends Kmeans and Spectral clusterings to
find clusters of performance from independent data points. On the contrary,
\toolname considers a model of programs where inputs are given as functional data.
Then, it uses non-parametric functional data clustering~\cite{jacques2014functional}
to detect clusters of performance. DPDEBUGGER~\cite{aaai18} is limited to linear
performance functions, while \toolname can model complex functions such as
polynomials.
Similar to~\cite{aaai18}, \toolname uses decision tree classifiers to pinpoint
code regions contributing to the differential performance.

\section{Threat to Validity}
\label{sec:threat}
\noindent \textbf{Evolutionary-based Fuzzing.}
Our approach requires a diverse set of inputs generated
automatically using the fuzzing component.
The quality of the debugging significantly depends on the characterization
of differential performance in the given input set.
Similar to existing evolutionary-based fuzzers,
our approach relies solely on heuristics to generate a diverse
set of inputs and is not guaranteed to find inputs that characterize
all performance classes.

\noindent \textbf{Benign Differential Performance vs Performance Bugs.}
In general, it is indeed challenging to determine whether a differential
performance is a bug or it is intrinsic to the problem being solved.
To mitigate this issue, we turn to
debugging with the help of auxiliary features from the
space of inputs and internals.

First, we find an explanation based on the input features
such as the type of solver.
Based on this, the user may decide the differences due
to using solver=`A’ versus solver=`B’ are benign whereas the differences
due to using tolerance=0.0 versus tolerance=0.000001 under the same
solver=`A’ are unexpected.

Once an unexpected behavior is detected, the next step is to investigate the
issue further in the source code and localize suspicious code regions for a fix.
These two steps help the user determine whether the differences are
intrinsic, or they are performance bugs that need to be fixed.

\noindent \textbf{Experiments and Comparisons to Existing Fuzzers.}
Existing approaches in fuzzing such as SlowFuzz~\cite{petsios2017slowfuzz}
and PerfFuzz~\cite{lemieux2018perffuzz} are mainly developed for C and C++
programs and extended recently for Java applications~\cite{DBLP:conf/icse/nilizadeh}.
Our work provides substantial (and unprecedented) support specifically for python-based
ML libraries.

To enable ourselves to compare \toolname against the existing
performance fuzzers, we adapt their main fuzzing algorithm and implement them in
our python-based fuzzing framework. This, however, can lead to degradations
in the performance of these fuzzers. To alleviate this in our comparisons,
we use the same functionality in all aspects of three fuzzers such as
mutation and crossover operations. The differences are in modeling population
(multiple populations with clustering versus single global population) and adding
new inputs to the population (based on the cost, the path, or combinations).
These differences are the specific choices of each fuzzer to achieve certain
goals as described in their algorithms.

\noindent \textbf{Overhead in Dynamic Analysis.}
We proposed a dynamic analysis approach to discover and understand
performance bugs.
Dynamic analysis often scales well to large applications.
However, as compared with static analysis, they present additional overheads
such as time required to discover variegated inputs and time needed for
data collection.

\noindent \textbf{Polynomial Functions and Decision Tree Models.}
Our design guiding principles are based on two important factors:
efficiency (especially for fuzzing) and human interpretability (for debugging).
For example, we restrict the search for performance functions to be polynomials
so as to generate inputs quickly.
Other models such as Gaussian processes can lead to better results, but they
may degrade the throughput of fuzzing.
Similarly, we use decision tree models to give interpretable explanations.
Graph models can be used to learn complex discriminants and overcome the decision
tree limitations such as the hyper-rectangular partitions of search spaces.
However, such models are notorious to be uninterpretable.

\noindent \textbf{Standard Algorithms as Benchmarks.}
We use standard sort, search, tree, and graph algorithms to evaluate our fuzzing.
While these algorithms do not contain performance bugs, they have (well-known)
diverse classes of performances, and finding those classes is a hard problem.
Characterizing these classes in the well-known algorithm through
fuzzing is important to manifest differential performance bugs in real-world
applications.

\noindent \textbf{Machine Learning Libraries as Case Studies.}
In this work, we focus on medium-sized ML libraries for few reasons.
First, there is a little bit of support for the performance aspects
of ML libraries.
Second, there is often a clear distinction between data
and parameters in ML libraries, and they tend to have
simpler performance functions such as linear or polynomial.
Our approach is applicable for general software given that there is
a clear measure defined to map inputs to the size.
Examples are the number of bytes in a file, the number of set bits in a key,
and the number of Kleene star in a regular expression.
Given this measure, our approach can characterize performance differences
and aid to localize the root causes. We left further analysis to apply
this technique on general and large software for future work.

\noindent \textbf{Time Measurements.}
We use actual execution times as opposed to abstractions such as the number
of executed lines in the case studies. While this is important to factor the
cost of black-box components (such as external libraries and solvers in other
languages) during the fuzzing, the noise in timing observations can lead to false
positive in the fuzzing process. To overcome this issue, we re-run the inputs
once more on NUC5i5RYH machine to allow for higher precisions.
To further mitigate the effects of environmental factors in NUC measurements,
we run the libraries in isolations and take the average of timing measurements
over multiple samples.

\section{Conclusion and Future Work}
\label{sec:conclusion}
We developed a method and a tool for differential performance analysis. We
showed that the fuzzing, clustering, and decision tree algorithms presented
for functional data are scalable to debug real-world machine learning libraries.
In addition, we illustrated the usefulness of our approach in finding multiple
performance bugs in these libraries and in comparing to existing performance
fuzzers.

For future work, there are few interesting directions.
One direction is to study security
implications of differential performance.
The feasibility of (hyper)parameter leaks~\cite{wang2018stealing}
via timing side channels in ML applications is a relevant and
challenging open problem. Another direction is to study
the relationships between accuracy and performance. Given
a lower-bound on the accuracy of a learning task, the idea is to
synthesize parameters and hyper-parameters in the model such that
the performance of underlying systems such as IoT and CPS
are optimized.

\begin{acks}
The authors would like to thank the anonymous reviewers for their comments
to improve this paper. In addition, the authors thank the developers of
\texttt{scikit-learn} for discussing the potential issues reported
by us. This work utilized resources from the University of Colorado Boulder
Research Computing Group, which is supported by NSF, CU Boulder, and CSU.
This research was supported by DARPA under agreement FA8750-15-2-0096.
\end{acks}

\bibliographystyle{ACM-Reference-Format}
\bibliography{papers}


\begin{thebibliography}{43}


\ifx \showCODEN    \undefined \def \showCODEN     #1{\unskip}     \fi
\ifx \showDOI      \undefined \def \showDOI       #1{#1}\fi
\ifx \showISBNx    \undefined \def \showISBNx     #1{\unskip}     \fi
\ifx \showISBNxiii \undefined \def \showISBNxiii  #1{\unskip}     \fi
\ifx \showISSN     \undefined \def \showISSN      #1{\unskip}     \fi
\ifx \showLCCN     \undefined \def \showLCCN      #1{\unskip}     \fi
\ifx \shownote     \undefined \def \shownote      #1{#1}          \fi
\ifx \showarticletitle \undefined \def \showarticletitle #1{#1}   \fi
\ifx \showURL      \undefined \def \showURL       {\relax}        \fi
\providecommand\bibfield[2]{#2}
\providecommand\bibinfo[2]{#2}
\providecommand\natexlab[1]{#1}
\providecommand\showeprint[2][]{arXiv:#2}

\bibitem[\protect\citeauthoryear{AFL}{AFL}{2016}]%
        {AFL}
\bibfield{author}{\bibinfo{person}{AFL}.} \bibinfo{year}{2016}\natexlab{}.
\newblock \bibinfo{title}{American fuzzy lop}.
\newblock \bibinfo{howpublished}{http://lcamtuf.coredump.cx/afl/}.
\newblock
\newblock
\shownote{Online.}


\bibitem[\protect\citeauthoryear{Breiman, Friedman, Stone, and Olshen}{Breiman
  et~al\mbox{.}}{1984}]%
        {breiman1984classification}
\bibfield{author}{\bibinfo{person}{Leo Breiman}, \bibinfo{person}{Jerome
  Friedman}, \bibinfo{person}{Charles~J Stone}, {and}
  \bibinfo{person}{Richard~A Olshen}.} \bibinfo{year}{1984}\natexlab{}.
\newblock \bibinfo{booktitle}{\emph{Classification and regression trees}}.
\newblock \bibinfo{publisher}{CRC press}.
\newblock


\bibitem[\protect\citeauthoryear{Chiba}{Chiba}{1998}]%
        {chiba1998javassist}
\bibfield{author}{\bibinfo{person}{Shigeru Chiba}.}
  \bibinfo{year}{1998}\natexlab{}.
\newblock \showarticletitle{Javassist - a reflection-based programming wizard
  for Java}. In \bibinfo{booktitle}{\emph{Proceedings of OOPSLA’98 Workshop
  on Reflective Programming in C++ and Java}}, Vol.~\bibinfo{volume}{174}.
\newblock


\bibitem[\protect\citeauthoryear{Goldsmith, Aiken, and Wilkerson}{Goldsmith
  et~al\mbox{.}}{2007}]%
        {goldsmith2007measuring}
\bibfield{author}{\bibinfo{person}{Simon~F Goldsmith}, \bibinfo{person}{Alex~S
  Aiken}, {and} \bibinfo{person}{Daniel~S Wilkerson}.}
  \bibinfo{year}{2007}\natexlab{}.
\newblock \showarticletitle{Measuring empirical computational complexity}. In
  \bibinfo{booktitle}{\emph{FSE}}. ACM, \bibinfo{pages}{395--404}.
\newblock


\bibitem[\protect\citeauthoryear{Hartmanis and Stearns}{Hartmanis and
  Stearns}{1965}]%
        {HS65}
\bibfield{author}{\bibinfo{person}{J. Hartmanis} {and} \bibinfo{person}{R.~E.
  Stearns}.} \bibinfo{year}{1965}\natexlab{}.
\newblock \showarticletitle{On the Computational Complexity of Algorithms}.
\newblock \bibinfo{journal}{\emph{Trans. Amer. Math. Soc.}}
  \bibinfo{volume}{117} (\bibinfo{year}{1965}), \bibinfo{pages}{285--306}.
\newblock
\showISSN{00029947}
\urldef\tempurl%
\url{http://www.jstor.org/stable/1994208}
\showURL{%
\tempurl}


\bibitem[\protect\citeauthoryear{Jacques and Preda}{Jacques and Preda}{2014}]%
        {jacques2014functional}
\bibfield{author}{\bibinfo{person}{Julien Jacques} {and}
  \bibinfo{person}{Cristian Preda}.} \bibinfo{year}{2014}\natexlab{}.
\newblock \showarticletitle{Functional data clustering: a survey}.
\newblock \bibinfo{journal}{\emph{Advances in Data Analysis and
  Classification}} \bibinfo{volume}{8}, \bibinfo{number}{3}
  (\bibinfo{year}{2014}), \bibinfo{pages}{231--255}.
\newblock


\bibitem[\protect\citeauthoryear{Kersten, Luckow, and
  P{\u{a}}s{\u{a}}reanu}{Kersten et~al\mbox{.}}{2017}]%
        {kersten2017poster}
\bibfield{author}{\bibinfo{person}{Rody Kersten}, \bibinfo{person}{Kasper
  Luckow}, {and} \bibinfo{person}{Corina~S P{\u{a}}s{\u{a}}reanu}.}
  \bibinfo{year}{2017}\natexlab{}.
\newblock \showarticletitle{POSTER: AFL-based Fuzzing for Java with Kelinci}.
  In \bibinfo{booktitle}{\emph{CCS}}. ACM, \bibinfo{pages}{2511--2513}.
\newblock


\bibitem[\protect\citeauthoryear{Lemieux, Padhye, Sen, and Song}{Lemieux
  et~al\mbox{.}}{2018}]%
        {lemieux2018perffuzz}
\bibfield{author}{\bibinfo{person}{Caroline Lemieux}, \bibinfo{person}{Rohan
  Padhye}, \bibinfo{person}{Koushik Sen}, {and} \bibinfo{person}{Dawn Song}.}
  \bibinfo{year}{2018}\natexlab{}.
\newblock \showarticletitle{Perffuzz: Automatically generating pathological
  inputs}. In \bibinfo{booktitle}{\emph{ISSTA}}. ACM,
  \bibinfo{pages}{254--265}.
\newblock


\bibitem[\protect\citeauthoryear{libFuzzer}{libFuzzer}{2016}]%
        {libFuzzer}
\bibfield{author}{\bibinfo{person}{libFuzzer}.}
  \bibinfo{year}{2016}\natexlab{}.
\newblock \bibinfo{title}{A library for coverage-guided fuzz testing (part of
  LLVM 3.9)}.
\newblock \bibinfo{howpublished}{http://llvm.org/docs/LibFuzzer.html}.
\newblock
\newblock
\shownote{Online.}


\bibitem[\protect\citeauthoryear{Lloyd}{Lloyd}{1982}]%
        {lloyd1982least}
\bibfield{author}{\bibinfo{person}{Stuart Lloyd}.}
  \bibinfo{year}{1982}\natexlab{}.
\newblock \showarticletitle{Least squares quantization in PCM}.
\newblock \bibinfo{journal}{\emph{IEEE transactions on information theory}}
  \bibinfo{volume}{28}, \bibinfo{number}{2} (\bibinfo{year}{1982}),
  \bibinfo{pages}{129--137}.
\newblock


\bibitem[\protect\citeauthoryear{Nilizadeh, Noller, and Pasareanu}{Nilizadeh
  et~al\mbox{.}}{2019}]%
        {DBLP:conf/icse/nilizadeh}
\bibfield{author}{\bibinfo{person}{Shirin Nilizadeh}, \bibinfo{person}{Yannic
  Noller}, {and} \bibinfo{person}{Corina~S. Pasareanu}.}
  \bibinfo{year}{2019}\natexlab{}.
\newblock \showarticletitle{DifFuzz: Differential Fuzzing for Side-Channel
  Analysis}.
\newblock \bibinfo{journal}{\emph{ICSE}} (\bibinfo{year}{2019}).
\newblock
\urldef\tempurl%
\url{http://arxiv.org/abs/1811.07005}
\showURL{%
\tempurl}


\bibitem[\protect\citeauthoryear{Oliphant}{Oliphant}{06  }]%
        {numpy}
\bibfield{author}{\bibinfo{person}{Travis Oliphant}.}
  \bibinfo{year}{2006--}\natexlab{}.
\newblock \bibinfo{title}{{NumPy}: A guide to {NumPy}}.
\newblock \bibinfo{howpublished}{http://www.numpy.org/}.
\newblock


\bibitem[\protect\citeauthoryear{Olivo, Dillig, and Lin}{Olivo
  et~al\mbox{.}}{2015}]%
        {olivo2015static}
\bibfield{author}{\bibinfo{person}{Oswaldo Olivo}, \bibinfo{person}{Isil
  Dillig}, {and} \bibinfo{person}{Calvin Lin}.}
  \bibinfo{year}{2015}\natexlab{}.
\newblock \showarticletitle{Static detection of asymptotic performance bugs in
  collection traversals}. In \bibinfo{booktitle}{\emph{PLDI}},
  Vol.~\bibinfo{volume}{50}. ACM, \bibinfo{pages}{369--378}.
\newblock


\bibitem[\protect\citeauthoryear{O'Whielacronx}{O'Whielacronx}{2018}]%
        {Trace}
\bibfield{author}{\bibinfo{person}{Zooko O'Whielacronx}.}
  \bibinfo{year}{2018}\natexlab{}.
\newblock \bibinfo{title}{A program/module to trace Python program or function
  execution}.
\newblock
  \bibinfo{howpublished}{https://github.com/python/cpython/blob/2.7/Lib/trace.py}.
\newblock
\newblock
\shownote{Online.}


\bibitem[\protect\citeauthoryear{Pedregosa, Varoquaux, Gramfort, Michel,
  Thirion, Grisel, Blondel, Prettenhofer, Weiss, Dubourg, Vanderplas, Passos,
  Cournapeau, Brucher, Perrot, and Duchesnay}{Pedregosa et~al\mbox{.}}{2011}]%
        {scikit-learn}
\bibfield{author}{\bibinfo{person}{F. Pedregosa}, \bibinfo{person}{G.
  Varoquaux}, \bibinfo{person}{A. Gramfort}, \bibinfo{person}{V. Michel},
  \bibinfo{person}{B. Thirion}, \bibinfo{person}{O. Grisel},
  \bibinfo{person}{M. Blondel}, \bibinfo{person}{P. Prettenhofer},
  \bibinfo{person}{R. Weiss}, \bibinfo{person}{V. Dubourg}, \bibinfo{person}{J.
  Vanderplas}, \bibinfo{person}{A. Passos}, \bibinfo{person}{D. Cournapeau},
  \bibinfo{person}{M. Brucher}, \bibinfo{person}{M. Perrot}, {and}
  \bibinfo{person}{E. Duchesnay}.} \bibinfo{year}{2011}\natexlab{}.
\newblock \showarticletitle{Scikit-learn: Machine Learning in {P}ython}.
\newblock \bibinfo{journal}{\emph{Journal of Machine Learning Research}}
  \bibinfo{volume}{12} (\bibinfo{year}{2011}), \bibinfo{pages}{2825--2830}.
\newblock


\bibitem[\protect\citeauthoryear{Petsios, Zhao, Keromytis, and Jana}{Petsios
  et~al\mbox{.}}{2017}]%
        {petsios2017slowfuzz}
\bibfield{author}{\bibinfo{person}{Theofilos Petsios}, \bibinfo{person}{Jason
  Zhao}, \bibinfo{person}{Angelos~D Keromytis}, {and} \bibinfo{person}{Suman
  Jana}.} \bibinfo{year}{2017}\natexlab{}.
\newblock \showarticletitle{Slowfuzz: Automated domain-independent detection of
  algorithmic complexity vulnerabilities}. In \bibinfo{booktitle}{\emph{CCS}}.
  ACM, \bibinfo{pages}{2155--2168}.
\newblock


\bibitem[\protect\citeauthoryear{Ramsay}{Ramsay}{2006}]%
        {ramsay2006functional}
\bibfield{author}{\bibinfo{person}{James~O Ramsay}.}
  \bibinfo{year}{2006}\natexlab{}.
\newblock \bibinfo{booktitle}{\emph{Functional data analysis}}.
\newblock \bibinfo{publisher}{Wiley Online Library}.
\newblock


\bibitem[\protect\citeauthoryear{Scikit-learn}{Scikit-learn}{2017}]%
        {rnd-forest-issue}
\bibfield{author}{\bibinfo{person}{Scikit-learn}.}
  \bibinfo{year}{2017}\natexlab{}.
\newblock \bibinfo{title}{Trees with MAE criterion are slow to train}.
\newblock
  \bibinfo{howpublished}{https://github.com/scikit-learn/scikit-learn/issues/9626}.
\newblock
\newblock
\shownote{Online.}


\bibitem[\protect\citeauthoryear{Scikit-learn}{Scikit-learn}{2018}]%
        {pair-wise-distance-issue}
\bibfield{author}{\bibinfo{person}{Scikit-learn}.}
  \bibinfo{year}{2018}\natexlab{}.
\newblock \bibinfo{title}{Sqeuclidean metric is much slower than euclidean}.
\newblock
  \bibinfo{howpublished}{https://github.com/scikit-learn/scikit-learn/issues/12600}.
\newblock
\newblock
\shownote{Online.}


\bibitem[\protect\citeauthoryear{Scikit-learn}{Scikit-learn}{2019a}]%
        {decision-tree-scikit-learn}
\bibfield{author}{\bibinfo{person}{Scikit-learn}.}
  \bibinfo{year}{2019}\natexlab{a}.
\newblock \bibinfo{title}{Decision Tree Classifier}.
\newblock
  \bibinfo{howpublished}{https://scikit-learn.org/stable/modules/tree.html}.
\newblock
\newblock
\shownote{Online.}


\bibitem[\protect\citeauthoryear{Scikit-learn}{Scikit-learn}{2019b}]%
        {gaussian-process-explanation}
\bibfield{author}{\bibinfo{person}{Scikit-learn}.}
  \bibinfo{year}{2019}\natexlab{b}.
\newblock \bibinfo{title}{Gaussian Process Classifier in scikit-learn:
  description}.
\newblock
  \bibinfo{howpublished}{https://scikit-learn.org/stable/modules/gaussian\_process.html\#gaussian-process-classification-gpc}.
\newblock
\newblock
\shownote{Online.}


\bibitem[\protect\citeauthoryear{Scikit-learn}{Scikit-learn}{2019c}]%
        {gaussian-process-implementations}
\bibfield{author}{\bibinfo{person}{Scikit-learn}.}
  \bibinfo{year}{2019}\natexlab{c}.
\newblock \bibinfo{title}{Gaussian Process Classifier in scikit-learn:
  implementations}.
\newblock
  \bibinfo{howpublished}{https://scikit-learn.org/stable/modules/generated/sklearn.gaussian\_process.GaussianProcess
  Classifier.html}.
\newblock
\newblock
\shownote{Online.}


\bibitem[\protect\citeauthoryear{Scikit-learn}{Scikit-learn}{2019d}]%
        {discriminant-analysis-lda-qda}
\bibfield{author}{\bibinfo{person}{Scikit-learn}.}
  \bibinfo{year}{2019}\natexlab{d}.
\newblock \bibinfo{title}{Linear and quadratic discriminant analysis}.
\newblock
  \bibinfo{howpublished}{https://scikit-learn.org/stable/modules/lda\_qda.html}.
\newblock
\newblock
\shownote{Online.}


\bibitem[\protect\citeauthoryear{Scikit-learn}{Scikit-learn}{2019e}]%
        {logistic-regression-explanation}
\bibfield{author}{\bibinfo{person}{Scikit-learn}.}
  \bibinfo{year}{2019}\natexlab{e}.
\newblock \bibinfo{title}{Logistic Regression in scikit-learn}.
\newblock
  \bibinfo{howpublished}{https://scikit-learn.org/stable/modules/linear\_model.html\#logistic-regression}.
\newblock
\newblock
\shownote{Online.}


\bibitem[\protect\citeauthoryear{Scikit-learn}{Scikit-learn}{2019f}]%
        {make-multilabel-classification}
\bibfield{author}{\bibinfo{person}{Scikit-learn}.}
  \bibinfo{year}{2019}\natexlab{f}.
\newblock \bibinfo{title}{Make Multilabel Classification}.
\newblock
  \bibinfo{howpublished}{sklearn.datasets.make\_multilabel\_classification}.
\newblock


\bibitem[\protect\citeauthoryear{Scikit-learn}{Scikit-learn}{2019g}]%
        {minibatch-kmeans}
\bibfield{author}{\bibinfo{person}{Scikit-learn}.}
  \bibinfo{year}{2019}\natexlab{g}.
\newblock \bibinfo{title}{Mini-batch KMeans}.
\newblock \bibinfo{howpublished}{sklearn.cluster.MiniBatchKMeans}.
\newblock


\bibitem[\protect\citeauthoryear{Scikit-learn}{Scikit-learn}{2019h}]%
        {logistic-regressio-issue}
\bibfield{author}{\bibinfo{person}{Scikit-learn}.}
  \bibinfo{year}{2019}\natexlab{h}.
\newblock \bibinfo{title}{Performance of Logistic Regression with saga}.
\newblock
  \bibinfo{howpublished}{https://github.com/scikit-learn/scikit-learn/issues/13316}.
\newblock
\newblock
\shownote{Online.}


\bibitem[\protect\citeauthoryear{Scikit-learn}{Scikit-learn}{2019i}]%
        {random-forest-regressor}
\bibfield{author}{\bibinfo{person}{Scikit-learn}.}
  \bibinfo{year}{2019}\natexlab{i}.
\newblock \bibinfo{title}{Random Forest Regressor}.
\newblock \bibinfo{howpublished}{sklearn.ensemble.RandomForestRegressor}.
\newblock


\bibitem[\protect\citeauthoryear{Scikit-learn}{Scikit-learn}{2020a}]%
        {logistic-regression-bug}
\bibfield{author}{\bibinfo{person}{Scikit-learn}.}
  \bibinfo{year}{2020}\natexlab{a}.
\newblock \bibinfo{title}{Performance bug in logistic regression with
  newton-cg}.
\newblock
  \bibinfo{howpublished}{https://github.com/scikit-learn/scikit-learn/issues/16186}.
\newblock
\newblock
\shownote{Online.}


\bibitem[\protect\citeauthoryear{Scikit-learn}{Scikit-learn}{2020b}]%
        {make-classification-dataset-bug}
\bibfield{author}{\bibinfo{person}{Scikit-learn}.}
  \bibinfo{year}{2020}\natexlab{b}.
\newblock \bibinfo{title}{Performance bug in Make Classification Data Set
  Util}.
\newblock
  \bibinfo{howpublished}{https://github.com/scikit-learn/scikit-learn/issues/16001}.
\newblock
\newblock
\shownote{Online.}


\bibitem[\protect\citeauthoryear{Scikit-learn}{Scikit-learn}{2020c}]%
        {make-classification-dataset-fix}
\bibfield{author}{\bibinfo{person}{Scikit-learn}.}
  \bibinfo{year}{2020}\natexlab{c}.
\newblock \bibinfo{title}{Performance bug in Make Classification Data Set Util
  fixed}.
\newblock
  \bibinfo{howpublished}{https://github.com/scikit-learn/scikit-learn/pull/16006/files}.
\newblock
\newblock
\shownote{Online.}


\bibitem[\protect\citeauthoryear{Scikit-learn}{Scikit-learn}{2020d}]%
        {logistic-regression-fix}
\bibfield{author}{\bibinfo{person}{Scikit-learn}.}
  \bibinfo{year}{2020}\natexlab{d}.
\newblock \bibinfo{title}{Performance bug in regression with newton-cg fixed}.
\newblock
  \bibinfo{howpublished}{https://github.com/scikit-learn/scikit-learn/pull/16266/files}.
\newblock
\newblock
\shownote{Online.}


\bibitem[\protect\citeauthoryear{Scikit-learn}{Scikit-learn}{2020e}]%
        {batch-generator-bug}
\bibfield{author}{\bibinfo{person}{Scikit-learn}.}
  \bibinfo{year}{2020}\natexlab{e}.
\newblock \bibinfo{title}{Performance bug in Util Batch Generator module}.
\newblock
  \bibinfo{howpublished}{https://github.com/scikit-learn/scikit-learn/issues/16158}.
\newblock
\newblock
\shownote{Online.}


\bibitem[\protect\citeauthoryear{Scikit-learn}{Scikit-learn}{2020f}]%
        {batch-generator-fix}
\bibfield{author}{\bibinfo{person}{Scikit-learn}.}
  \bibinfo{year}{2020}\natexlab{f}.
\newblock \bibinfo{title}{Performance bug in Util Batch Generator module
  fixed}.
\newblock
  \bibinfo{howpublished}{https://github.com/scikit-learn/scikit-learn/pull/16181/files}.
\newblock
\newblock
\shownote{Online.}


\bibitem[\protect\citeauthoryear{Sedgewick and Wayne}{Sedgewick and
  Wayne}{2011}]%
        {algorithm-book}
\bibfield{author}{\bibinfo{person}{Robert Sedgewick} {and}
  \bibinfo{person}{Kevin Wayne}.} \bibinfo{year}{2011}\natexlab{}.
\newblock \bibinfo{title}{Algorithms (4th ed.)}.
\newblock
\newblock
\newblock
\shownote{Addison-Wesley Professional.}


\bibitem[\protect\citeauthoryear{Song and Lu}{Song and Lu}{2014}]%
        {song2014statistical}
\bibfield{author}{\bibinfo{person}{Linhai Song} {and} \bibinfo{person}{Shan
  Lu}.} \bibinfo{year}{2014}\natexlab{}.
\newblock \showarticletitle{Statistical debugging for real-world performance
  problems}.
\newblock \bibinfo{journal}{\emph{OOPSLA}} \bibinfo{volume}{49},
  \bibinfo{number}{10} (\bibinfo{year}{2014}), \bibinfo{pages}{561--578}.
\newblock


\bibitem[\protect\citeauthoryear{Tensorflow}{Tensorflow}{2019}]%
        {train-slow-cpu-issue}
\bibfield{author}{\bibinfo{person}{Tensorflow}.}
  \bibinfo{year}{2019}\natexlab{}.
\newblock \bibinfo{title}{Transpose can be very slow on CPU}.
\newblock
  \bibinfo{howpublished}{https://github.com/tensorflow/tensorflow/issues/27383}.
\newblock
\newblock
\shownote{Online.}


\bibitem[\protect\citeauthoryear{Tizpaz{-}Niari, {\v C}ern\'y, Chang, and
  Trivedi}{Tizpaz{-}Niari et~al\mbox{.}}{2018}]%
        {aaai18}
\bibfield{author}{\bibinfo{person}{Saeid Tizpaz{-}Niari},
  \bibinfo{person}{Pavol {\v C}ern\'y}, \bibinfo{person}{Bor{-}Yuh~Evan Chang},
  {and} \bibinfo{person}{Ashutosh Trivedi}.} \bibinfo{year}{2018}\natexlab{}.
\newblock \showarticletitle{Differential Performance Debugging with
  Discriminant Regression Trees}. In \bibinfo{booktitle}{\emph{AAAI}}.
  \bibinfo{pages}{2468--2475}.
\newblock


\bibitem[\protect\citeauthoryear{Tizpaz-Niari, {\v{C}}ern{\`y}, Chang,
  Sankaranarayanan, and Trivedi}{Tizpaz-Niari et~al\mbox{.}}{2017}]%
        {tizpaz2017discriminating}
\bibfield{author}{\bibinfo{person}{Saeid Tizpaz-Niari}, \bibinfo{person}{Pavol
  {\v{C}}ern{\`y}}, \bibinfo{person}{Bor-Yuh~Evan Chang},
  \bibinfo{person}{Sriram Sankaranarayanan}, {and} \bibinfo{person}{Ashutosh
  Trivedi}.} \bibinfo{year}{2017}\natexlab{}.
\newblock \showarticletitle{Discriminating Traces with Time}. In
  \bibinfo{booktitle}{\emph{TACAS}}. Springer, \bibinfo{pages}{21--37}.
\newblock


\bibitem[\protect\citeauthoryear{Tizpaz-Niari, Cerny, and Trivedi}{Tizpaz-Niari
  et~al\mbox{.}}{2020}]%
        {FuncSideChan18}
\bibfield{author}{\bibinfo{person}{Saeid Tizpaz-Niari}, \bibinfo{person}{Pavol
  Cerny}, {and} \bibinfo{person}{Ashutosh Trivedi}.}
  \bibinfo{year}{2020}\natexlab{}.
\newblock \bibinfo{title}{Data-Driven Debugging for Functional Side Channels}.
\newblock \bibinfo{howpublished}{\url{https://arxiv.org/abs/1808.10502}}.
\newblock
\newblock
\shownote{In NDSS.}


\bibitem[\protect\citeauthoryear{Wang and Gong}{Wang and Gong}{2018}]%
        {wang2018stealing}
\bibfield{author}{\bibinfo{person}{Binghui Wang} {and}
  \bibinfo{person}{Neil~Zhenqiang Gong}.} \bibinfo{year}{2018}\natexlab{}.
\newblock \showarticletitle{Stealing hyperparameters in machine learning}. In
  \bibinfo{booktitle}{\emph{2018 IEEE Symposium on Security and Privacy (SP)}}.
  IEEE, \bibinfo{pages}{36--52}.
\newblock


\bibitem[\protect\citeauthoryear{Wong, Gao, Li, Abreu, and Wotawa}{Wong
  et~al\mbox{.}}{2016}]%
        {wong2016survey}
\bibfield{author}{\bibinfo{person}{W~Eric Wong}, \bibinfo{person}{Ruizhi Gao},
  \bibinfo{person}{Yihao Li}, \bibinfo{person}{Rui Abreu}, {and}
  \bibinfo{person}{Franz Wotawa}.} \bibinfo{year}{2016}\natexlab{}.
\newblock \showarticletitle{A survey on software fault localization}.
\newblock \bibinfo{journal}{\emph{IEEE Transactions on Software Engineering}}
  \bibinfo{volume}{42}, \bibinfo{number}{8} (\bibinfo{year}{2016}),
  \bibinfo{pages}{707--740}.
\newblock


\bibitem[\protect\citeauthoryear{Zeller, Gopinath, B{\"o}hme, Fraser, and
  Holler}{Zeller et~al\mbox{.}}{2019}]%
        {fuzzingbook2019:index}
\bibfield{author}{\bibinfo{person}{Andreas Zeller}, \bibinfo{person}{Rahul
  Gopinath}, \bibinfo{person}{Marcel B{\"o}hme}, \bibinfo{person}{Gordon
  Fraser}, {and} \bibinfo{person}{Christian Holler}.}
  \bibinfo{year}{2019}\natexlab{}.
\newblock \showarticletitle{The Fuzzing Book}.
\newblock In \bibinfo{booktitle}{\emph{The Fuzzing Book}}.
  \bibinfo{publisher}{Saarland University}.
\newblock
\urldef\tempurl%
\url{https://www.fuzzingbook.org/}
\showURL{%
\tempurl}
\newblock
\shownote{Retrieved 2019-09-09 16:42:54+02:00.}


\end{thebibliography}

\end{document}